\documentclass{article} % For LaTeX2e
\usepackage{iclr2027_conference,times}
\usepackage{bm}
\usepackage{colortbl}
\usepackage{placeins}
\usepackage{capt-of}
\definecolor{GroupGray}{HTML}{F2F2F5}
\usepackage{booktabs}
\usepackage{xcolor}

\usepackage{soul}
\definecolor{RevisionYellow}{HTML}{FFF2A8}
\sethlcolor{RevisionYellow}

\definecolor{BestRed}{HTML}{B42318}
\definecolor{SecondPurple}{HTML}{5B3F99}
\usepackage{multirow}

\usepackage{amsmath,amsfonts,bm}

\def\eqref#1{equation~\ref{#1}}
\def\1{\bm{1}}

\DeclareMathAlphabet{\mathsfit}{\encodingdefault}{\sfdefault}{m}{sl}
\SetMathAlphabet{\mathsfit}{bold}{\encodingdefault}{\sfdefault}{bx}{n}

\usepackage{amsmath}
\usepackage{amssymb}
\usepackage{hyperref}
\usepackage{url}

\title{DeepQuest: From Knowledge Gaps to Scientific Discovery in Alzheimer's Disease}

\makeatletter
\newcommand{\affilthanks}[2]{%
  \begingroup
  \def\@makefnmark{\textsuperscript{#1,\@thefnmark}}%
  \thanks{#2}%
  \endgroup
}
\makeatother

\author{%
  {\bfseries
    Chunbo Jiang\affilthanks{1}{Equal contribution.},
    Yifei Chen\textsuperscript{2,*},
    Beining Wu\textsuperscript{1},
    Fuyou Mao\textsuperscript{3},
    Weihong Chen\textsuperscript{1},%
  }\\
  {\bfseries
    Enyu Bao\textsuperscript{1},
    Mingxuan Liu\textsuperscript{2},
    Feiwei Qin\affilthanks{1}{Corresponding authors: Feiwei Qin and Qiyuan Tian.},
    Qiyuan Tian\textsuperscript{2,\textdagger}%
  }\\
  {\normalfont
    \textsuperscript{1}Hangzhou Dianzi University\quad
    \textsuperscript{2}Tsinghua University\quad
    \textsuperscript{3}Central South University%
  }%
}

\iclrfinalcopy % Uncomment for camera-ready version, but NOT for submission.
\usepackage{graphicx}

\begin{document}

\maketitle

\begin{abstract}
Automating experiments does not necessarily advance scientific understanding. Research agents must identify meaningful knowledge gaps, determine what available data can resolve, and assess whether their findings advance existing knowledge. We introduce DeepQuest, a knowledge-gap-driven AI scientist framework for Alzheimer’s disease research. DeepQuest transforms unresolved questions from domain reviews into traceable, testable hypotheses and prioritizes them through data feasibility screening and groupwise Elo ranking. It couples iterative experimentation with two-level auditing that separates experimental reliability from knowledge contribution relative to two frozen literature corpora. On OASIS-3, DeepQuest produces 42 feasible hypotheses and outperforms the strongest baselines in 14 of 15 evaluator–dimension comparisons across three language-model reviewers. The full framework achieves an audit-assessed scientific finding rate of 95.24\%, outperforming both general-purpose executors and domain-specific research agents. This scientific finding rate remains stable across literature corpora and evaluator model families. Representative findings concern neuropsychiatric symptoms and dementia risk, longitudinal amyloid spatial stability, and neocortical tau-associated clinical impairment, with key associations supported in independent ADNI and AIBL cohorts. These results show how aligning knowledge gaps, data feasibility, and evidence auditing can turn automated research into traceable, evidence-supported scientific discovery. Our source code is
available at \url{https://anonymous.4open.science/r/DeepQuest-3066/}.
\end{abstract}

\begin{figure}[ht]
    \centering
    \includegraphics[width=\linewidth]{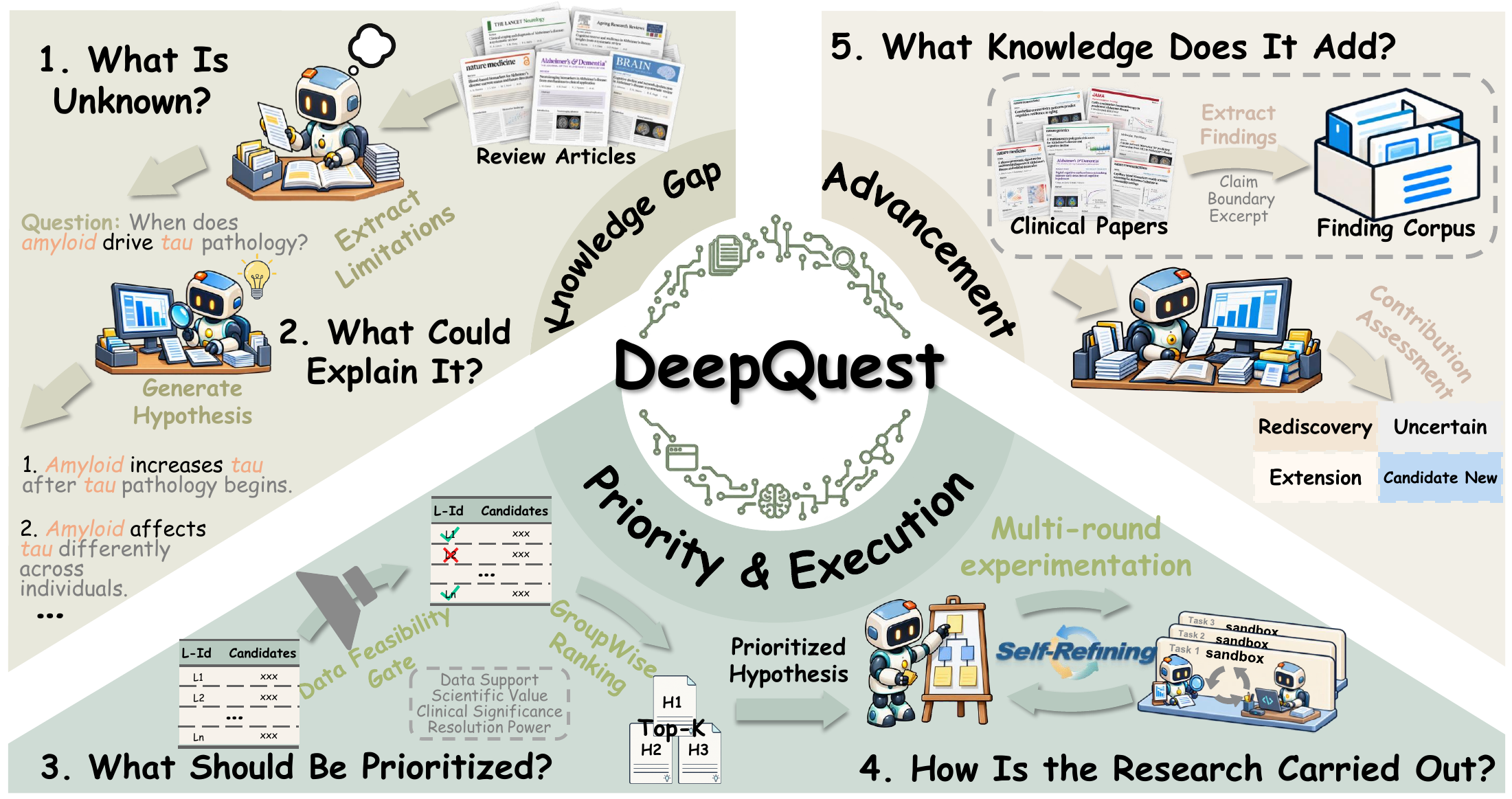}
    \caption{Overview of the DeepQuest framework for automated AD research. The framework integrates knowledge-gap-driven hypothesis formulation, data-constrained task selection, iterative experimentation, and literature-based knowledge contribution assessment.}
    \label{fig:overview}
\end{figure}
%\newpage
\section{Introduction}

A central challenge in Alzheimer’s disease (AD) research is to translate the growing body of heterogeneous evidence into testable scientific questions that advance understanding of the disease. The complex pathology of AD spans multiple biological levels \citep{jack2024revised,zhao2025multiform,shwab2025leveraging}. Large scale cohorts and data sharing initiatives have assembled multimodal data encompassing clinical assessments, cognitive measures, genetic information, biofluid biomarkers, magnetic resonance imaging (MRI), and positron emission tomography (PET), providing a foundation for investigating disease states, progression, and pathological associations \citep{travaglia2024data,toga2024informatics,veitch2026contributions,weiner2025overview}. DiaMond learns cross-modal disease representations from MRI and PET \citep{liyitong2025}. CARE-AD extracts information on disease evolution from longitudinal clinical records \citep{li2025care}. The multitask model proposed by \citet{wang2025deep} jointly analyzes brain age, cognition, and amyloid pathology. These advances strengthen data analysis while placing greater demands on evidence organization, research question formulation, and the evaluation of existing disease understanding.

Research agents have extended automation across multiple stages of the scientific workflow, yet completing the workflow does not by itself ensure scientifically valuable discoveries. General-purpose AI Scientist systems such as DeepScientist can generate candidate hypotheses, plan experiments, and iteratively refine their exploration based on feedback \mbox{\citep{weng2026deepscientist,yamada2025ai,tang2026ai}}. Domain-specific systems, such as Medical AI Scientist, AMID and SPARK, further integrate domain knowledge, specialized tools, and quality control mechanisms to support complex analytical workflows in medicine and neuroscience \citep{wu2026towards,liu2026towards,han2026towards,trost2026agentic}. However, generating hypotheses and executing experiments for a predefined objective still requires assessing whether they address important knowledge gaps, are testable with available data, and advance existing understanding.

Autonomous scientific discovery in AD requires three interdependent decisions: problem formulation, task selection, and knowledge contribution assessment. First, limitations identified in the literature must be translated into experimentally testable hypotheses with explicit evidential provenance, thereby establishing a traceable link between existing knowledge and newly formulated research questions. Second, scientifically meaningful questions are not necessarily answerable with available medical data. Task selection must therefore balance scientific value against execution cost under constraints imposed by data feasibility and experimental resources. Third, experimental reliability and scientific contribution should be assessed separately. Improvements in predictive performance or statistical significance may support specific experimental conclusions, but neither alone establishes that a study has produced new understanding of the disease. Integrating these stages enables research agents to pursue questions according to their potential knowledge gain.

To this end, we propose DeepQuest, an AI scientist framework for AD research designed to advance existing knowledge by integrating knowledge gap identification, data-constrained task selection, experimental validation, and knowledge contribution assessment as shown in Fig.~\ref{fig:overview}. DeepQuest first extracts structured research limitations from domain reviews and transforms them into testable hypotheses. It then applies multi-stage screening and relative priority ranking under data feasibility constraints to select research tasks that are both scientifically valuable and feasible to execute. During experimentation, DeepQuest combines iterative execution with a two-level audit of workflow reliability and the contribution of conclusions beyond existing knowledge. This integrates problem formulation, task selection, and conclusion assessment into a unified research process.

Experimental results support the effectiveness of DeepQuest at three levels: hypothesis quality, knowledge contribution assessment, and validation on independent cohorts. Across all three evaluator models, mean evaluation scores for the 42 executable hypotheses generated by DeepQuest were better than those for five strong generative-model baselines. The evidence audit classified the resulting experimental claims from 40 of 42 hypotheses (95.24\%) as Candidate New Discovery or Extension. All ten top-ranked hypotheses were identified as having the potential to advance existing knowledge. Three representative case studies on amyloid spatial topology, neuropsychiatric symptoms and dementia risk, and the relationship between tau pathology and clinical impairment further demonstrate DeepQuest’s ability to carry research from problem formulation through evidence validation. The corresponding findings were broadly supported in independent external cohorts.

Overall, our contributions are as follows:
\begin{itemize}
\item We introduce a review-driven approach to formulating scientific questions in AD research. The method transforms research limitations identified in AD domain reviews into structured representations and testable hypotheses, establishing a traceable mapping from literature evidence and knowledge gaps to research questions.
\item We introduce a data-constrained task selection and optimization approach. It formulates task selection as an optimization problem over the intersection between the literature-derived research space and the data feasibility region. Through multi-stage screening and priority ranking, it balances scientific value with execution feasibility.
\item We develop a two-level auditing mechanism that separates experimental reliability from knowledge contribution. DeepQuest independently examines whether experiments are reliable and conclusions extend beyond existing knowledge, providing traceable evidence for assessing candidate scientific discoveries and their contributions.

\end{itemize}
%% Please note that we have introduced automatic line number generation
%% into the style file for \LaTeXe. This is to help reviewers
%% refer to specific lines of the paper when they make their comments. Please do
%% NOT refer to these line numbers in your paper as they will be removed from the
%% style file for the final version of accepted papers.

\section{Related Work}

\textbf{General Autonomous Scientific Research Systems.}  General-purpose AI research systems automate multiple research stages by connecting hypothesis generation, experimental search, and feedback \citep{lu2024ai,schmidgall2025agent,mitchener2025kosmos}. AI Scientist-v2 \citep{yamada2025ai} organizes experimental exploration through agent-based tree search and performs result analysis and manuscript generation. AI-Researcher \citep{tang2026ai} integrates literature investigation, hypothesis formulation, and algorithm implementation through multiagent collaboration. DeepScientist \citep{weng2026deepscientist} formulates scientific discovery as a Bayesian optimization problem and combines discovery memory with hierarchical validation to iteratively select and explore promising candidate directions. Collectively, these systems demonstrate the capacity for open-ended scientific ideation and autonomous research direction selection, with representative evaluations conducted primarily on machine learning tasks. Applying these mechanisms to AD cohort studies requires aligning hypotheses with target populations, observable variables, follow-up windows, and confounding-control requirements, so  empirical evidence constrains question selection.

\textbf{Biomedical Research Agents.}  Biomedical research agents advance hypothesis exploration for medical problems by incorporating domain knowledge and specialized validation procedures \citep{huang2026autonomous}. Co-Scientist \citep{gottweis2026accelerating} iteratively refines research directions through hypothesis generation, critique, and ranking, and conducts experimental validation in biomedical settings. Robin \citep{ghareeb2026multi} combines literature-based reasoning, candidate selection, and experimental feedback to support the discovery of therapeutic candidates. Medical AI Scientist \citep{wu2026towards} provides three modes for paper-based reproduction, literature-inspired innovation, and task-driven exploration, using coordinated medical and engineering reasoning to identify research gaps, formulate hypotheses, and execute experiments. These systems integrate literature-grounded reasoning and question exploration into scientific workflows, providing important points of reference for developing autonomous biomedical research.

Research agents for medical imaging and clinical cohort studies further incorporate data adaptation, analytical reliability, and hypothesis validation into the scientific workflow. AMID \citep{liu2026towards} plans candidate methods according to the characteristics and constraints of available data and integrates experimental auditing into model optimization. NEXUS \citep{han2026towards} uses quality feedback to revise neuroimaging analysis pipelines. NeuroAgent \citep{zhong2026neuroagent} connects multimodal preprocessing, predictive modeling, and interactive analysis within a unified workflow. More closely related to our setting, the Alzheimer’s Disease and Related Dementias (ADRD) hypothesis-testing system proposed by \citet{saichandran2025agentic} maps literature-derived hypotheses to available variables in the NACC cohort and uses iterative analysis and methodological review to assess testability and conduct statistical tests. Its evaluation focuses primarily on replicating previously reported associations. Our work extends the focus to the formulation, selection, and contribution assessment of candidate research questions in AD. DeepQuest constructs structured domain-specific knowledge gaps from domain reviews, prioritizes research tasks under data constraints, and separately audits experimental reliability and the incremental contribution of conclusions beyond existing knowledge.

\begin{figure}[ht]
    \centering
    \includegraphics[width=\linewidth]{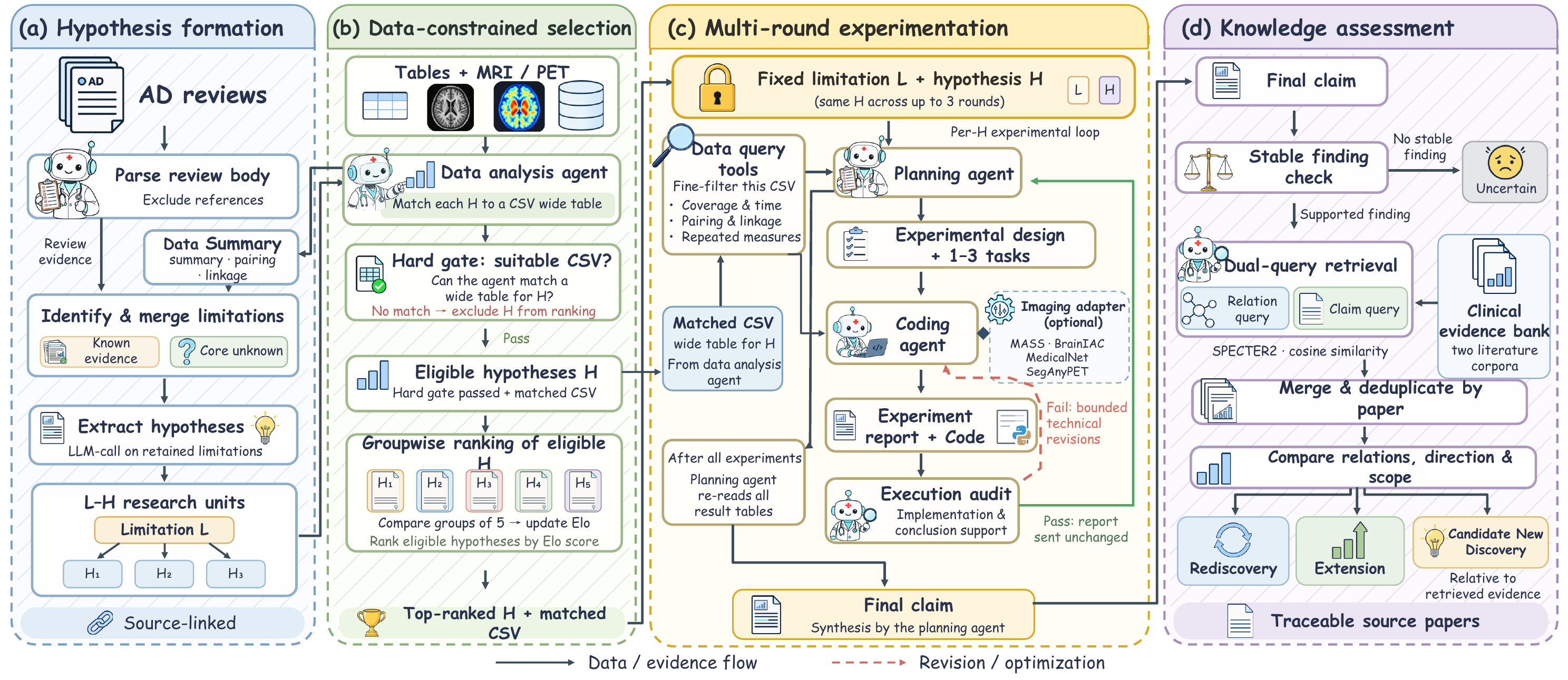}
    \caption{Architecture of DeepQuest. The framework consists of four stages: (a) knowledge-gap-driven hypothesis formulation, (b) data-constrained task selection, (c) iterative experimentation with execution auditing, and (d) literature-based knowledge contribution assessment.}
    \label{fig:architecture}
\end{figure}

\section{Method}

\subsection{Problem Formulation and Framework Overview}

We formulate automated AD research as a knowledge-gap-driven
process of task selection under data constraints and evidence
assessment. Given a collection of domain reviews $R$ and  medical
data $D$, DeepQuest constructs candidate research directions
$u_i = (l_i, H_i)$, where $l_i$ denotes a structured research
limitation and $H_i$ denotes the corresponding set of hypotheses.
It then proceeds through knowledge gap extraction, data feasibility
assessment, research priority ranking, iterative experimentation,
and knowledge contribution assessment, as illustrated in Fig.~\ref{fig:architecture}.

\subsection{Review-driven Knowledge Gap and Hypothesis Construction}

Knowledge gap extraction organizes established knowledge and key unknowns from domain reviews into traceable research limitations. DeepQuest extracts findings and unresolved relationships concerning AD pathology, biomarkers, imaging, and clinical research. Each limitation records known facts, unresolved phenomena, and supporting source evidence. Semantically equivalent unknowns across reviews are consolidated into a unified limitation $l_i$.

Hypothesis construction transforms research limitations into scientific propositions testable against observable outcomes. For each $l_i$, DeepQuest generates a hypothesis set  $H_i=\{h_{ij}\}_{j=1}^{n_i}$ and specifies empirical predictions that distinguish competing explanations. Each knowledge gap and its hypothesis set constitute a candidate research direction. This organization separates unresolved questions from explanations while preserving traceability to the original problem.

\subsection{Data Feasibility Assessment and Research Asset Preparation}

The data summary provides an initial assessment of whether candidate research questions can be examined with the available data. A data analysis agent examines the available variables, subject structure, imaging records, and multimodal longitudinal follow-up relationships, and constructs the following comprehensive structured representation:

\begin{equation}
z_D = A_{\text{data}}(D)
= \left(
z_{\text{summary}},
z_{\text{pairing}},
z_{\text{linkage}}
\right)
\label{eq:data_summary},
\end{equation}
where $z_{\text{summary}}$ summarizes variable semantics and measurement information. $z_{\text{pairing}}$ characterizes joint observation patterns across subjects, visits, and temporal dimensions. $z_{\text{linkage}}$ records entity identifiers, temporal keys, and cross-table linkage relationships. Based on this structured summary, DeepQuest applies soft gating to remove research directions that are clearly unsupported by the available data, according to variable semantics and observational coverage.

Hypothesis-level assessment determines whether available data meet the requirements for empirical validation. A data preparation agent examines each hypothesis alongside the structured summary and raw data, checking correspondence among variables, samples, temporal information, and imaging modalities. The feasible hypothesis set for each knowledge gap is denoted by $H_i^{\mathrm{feas}}$, and the retained individual hypotheses form $U_D$. This stage establishes testability and specific analytical strategies are determined during experimentation.

Research asset preparation organizes the validated data into reusable inputs for downstream experiments. DeepQuest registers verified assets according to data granularity, variable type, and scope of applicability, and prioritizes retrieval and reuse of existing results. For data that have not yet been curated, the system extracts and integrates the required information according to the variables and observation levels specified by each hypothesis. Imaging preprocessing is performed in a modality-specific manner. Structured data and preprocessed images are available for later experiments.

\subsection{Groupwise Comparative Research Prioritization}

Research ranking compares the value of exploring different knowledge gaps among data feasible candidates under the available cohort and modality constraints. DeepQuest further filters hypotheses that address core unresolved questions in AD, offer potential knowledge gain, and admit empirically discriminative predictions, yielding the candidate set $U_C \subseteq U_D$.

Global research priorities are established through repeated comparisons within groups of five candidates. Inspired by Co-Scientist’s pairwise ranking strategy \citep{gottweis2026accelerating}, DeepQuest evaluates each group using anonymized identifiers and randomized presentation order. Let $G_t$ denote the set of candidate indices in round $t$ , with $|G_t|=5$. Let $s_i^{(t)}$ denote the current score of candidate $i$, and let $\rho_i^{(t)} \in \{1,\ldots,5\}$ denote its rank within the group. For any two candidates $i,j$ in $G_t$, we compute the expected pairwise probability that candidate $i$ is preferred over candidate $j$ using the Elo rating model \citep{elo1978rating}:
\begin{equation}
p_{ij}^{(t)}
=
\frac{1}{
1 + 10^{\left(s_j^{(t)} - s_i^{(t)}\right)/400}
}
\label{eq:pairwise_probability}.
\end{equation}
The scores are synchronously updated according to the within-group rankings as follows:
\begin{equation}
s_i^{(t+1)}
=
s_i^{(t)}
+
\eta
\left[
\frac{5-\rho_i^{(t)}}{4}
-
\frac{1}{4}
\sum_{j \in G_t \setminus \{i\}}
p_{ij}^{(t)}
\right],
\quad i \in G_t
\label{eq:score_update},
\end{equation}
where $\eta$ denotes the update step size. The first term represents the proportion of the other four candidates that a given candidate outranks, while the second term is the average expected pairwise preference probability computed from the scores before the update. After $T_{\mathrm{rank}}$ comparison rounds, DeepQuest prioritizes individual hypotheses by their final scores, which indicate relative research priority only.

\subsection{Iterative Experimentation and Execution Auditing}

Experimental planning translates a selected hypothesis into an analytical task with explicit comparators and a defined evidence pathway. For a hypothesis $h_k$ entering the experimental stage, the planning agent fixes its associated research limitation and core proposition, invokes data tools to verify the data structures required for the current round, and constructs the shared dataset $D_t$. Each round consists of an experimental design $P_t$ and an execution task $T_t$, which specify the analysis target, variable configuration, statistical method, and expected evidence. Given $P_t$, $T_t$ and the available dataset $D_t$, the result of round $t$ is expressed as
\begin{equation}
R_t = f_{\mathrm{exec}}(P_t, T_t, D_t)
\label{eq:execution_result},
\end{equation}
where $f_{\mathrm{exec}}$ denotes the automated experimental execution procedure and $R_t$ denotes the resulting evidence from the current experimental round.

Experimental execution and technical auditing check whether the implementation follows the current design and whether the results support the reported conclusions. A coding agent performs data loading, sample construction, variable processing, and statistical or machine learning analysis in an isolated sandbox. For imaging tasks, representations are extracted from predeployed models through a unified adapter. The audit agent inspects the code implementation, core comparisons, and the correspondence between evidence and conclusions under the current design, and provides technical revision feedback when necessary. The experimental design remains fixed throughout revision. If the analysis still fails to pass the audit after reaching the maximum allowed number of revisions, the run is marked as an irrecoverable execution failure.

Cross-round optimization updates the experimental plan to address unresolved evidence gaps while keeping the core hypothesis unchanged. At the end of each round, the planning agent identifies remaining evidence gaps from the current result $R_t$ and determines whether to revise the comparator, data organization, or analytical design based on audit feedback:
\begin{equation}
P_{t+1} = f_{\mathrm{update}}(h_k, P_t, R_t),
\label{eq:plan_update}
\end{equation}
where $f_{\mathrm{update}}$ denotes the cross-round experimental design update process. DeepQuest conducts at most three rounds, then summarizes the conclusions and their overall scope of applicability under the data constraints. Within-round auditing checks implementation–design consistency, while cross-round planning develops the evidence around the original hypothesis.

\subsection{Literature-Based Assessment of Knowledge Contribution}

The literature evidence repository provides a traceable reference for comparing experimental conclusions against prior research.We decompose clinical studies on AD and related topics into individual finding entries to construct the evidence repository $K$ where each entry $e \in K$ records the core claim, boundary and excerpt. This repository defines the literature scope against which subsequent knowledge contribution is assessed.

Evidence sufficiency is assessed before literature comparison to distinguish interpretable conclusions from insufficiently supported results. Results are labeled as Uncertain when valid evidence is lacking because of inadequate statistical power, unstable estimates, nonestimable parameters, or incomplete analysis. Conclusions supported by evidence, including directional associations, conditional differences, or null effects, proceed to literature comparison. Statistical nonsignificance alone is not treated as evidence of a null effect in isolation.

Literature-grounded scientific synthesis and novelty assessment increasingly combine broad retrieval, semantic reranking, citation verification, and structured comparison against prior work \citep{lin2025evaluating,shahid2025literature}. Dual-route retrieval matches both the research relationship and the specific claim to obtain complementary literature evidence. For a claim $c$ , DeepQuest constructs a relation query $q_{\mathrm{rel}}(c)$ and a claim query $q_{\mathrm{claim}}(c)$. SPECTER2 \citep{singh2023scirepeval} is used to encode both queries and literature entries into vector representations, which are ranked by cosine similarity:
\begin{equation}
\operatorname{sim}(q,e)
=
\frac{
f_Q(q)^{\top} f_E(e)
}{
\left\| f_Q(q) \right\|_2
\left\| f_E(e) \right\|_2
}
\label{eq:similarity}.
\end{equation}
Each route’s results are deduplicated at the paper level. The top 16 papers from each route are retained and merged into a candidate evidence set covering at most 32 unique papers.

Knowledge relation auditing assesses each experimental claim against the retrieved evidence based on relation structure, claim direction, and scope. Claims consistent with existing evidence and introducing no new proposition are labeled Rediscovery. Those revealing additional temporal, interaction, nonlinear, spatial, or population-specific patterns are labeled Extension. Claims are labeled Candidate New Discovery when the retrieved evidence contains no direct precedent for the same underlying relationship or only broadly related precedents. DeepQuest retains the supporting rationale and source literature. Candidate New Discovery denotes potential knowledge gain relative to the retrieved evidence from the frozen repository, without establishing the absence of precedents in the broader literature. Execution auditing assesses procedural consistency, whereas knowledge relation auditing assesses literature level contribution.

\section{Experiment}

\subsection{ Experimental Setup and Evaluation Protocol}

We evaluate DeepQuest at three levels: hypothesis quality, research output, and medical discovery. OASIS-3 \citep{lamontagne2019oasis} serves as the primary cohort. Three case studies are selected to provide complementary clinical and scientific coverage, and their analyses are further validated on independent ADNI \citep{mueller2005ways} or AIBL \citep{ellis2009australian} cohorts to assess transferability.

We compare hypotheses generated by DeepQuest against five language model baselines All methods generate hypotheses from the same review set. Three independent language model based reviewers evaluate each hypothesis along five dimensions: novelty, scientific advancement, clinical significance, empirical discriminative power, and feasibility.

We include Codex \mbox{\citep{openai2025codex}} and Claude Code \mbox{\citep {anthropic2025claudecode}} as general-purpose baselines, and DiscoADRD \mbox{\citep{saichandran2025agentic}} and Biomni \mbox{\citep{huang2026autonomous}} as domain-specific baselines. Codex and Claude Code each execute three hypothesis sets generated by GPT models (GPT-H), Claude models (Claude-H), and DeepQuest (DeepQuest-H). Research outcomes are assessed against a frozen corpus of 200 clinical studies on AD and related topics published after February 16, 2026. This date marks GPT-5.6’s reported knowledge cutoff and is later than those of the Claude models used in this study. We further evaluate the outcomes against an expanded frozen corpus of 4604 articles from 2024 to 2026, examining the stability of scientific finding rates and contribution labels under broader literature coverage.

We define the Scientific Finding Rate (SFR) as the proportion of evaluated tasks classified by the audit as Candidate New Discovery, Rediscovery, or Extension. Tasks that fail to produce a result because of execution failure, timeout, or reaching the maximum number of audit revisions are labeled Skip. We further conduct component-wise ablation studies to assess the contributions of individual modules. Further implementation details are provided in Appendix ~\ref{app:A}.

\subsection{Hypothesis Quality and Research Prioritization}

\begin{figure}[ht]
    \centering
    \includegraphics[width=\linewidth]{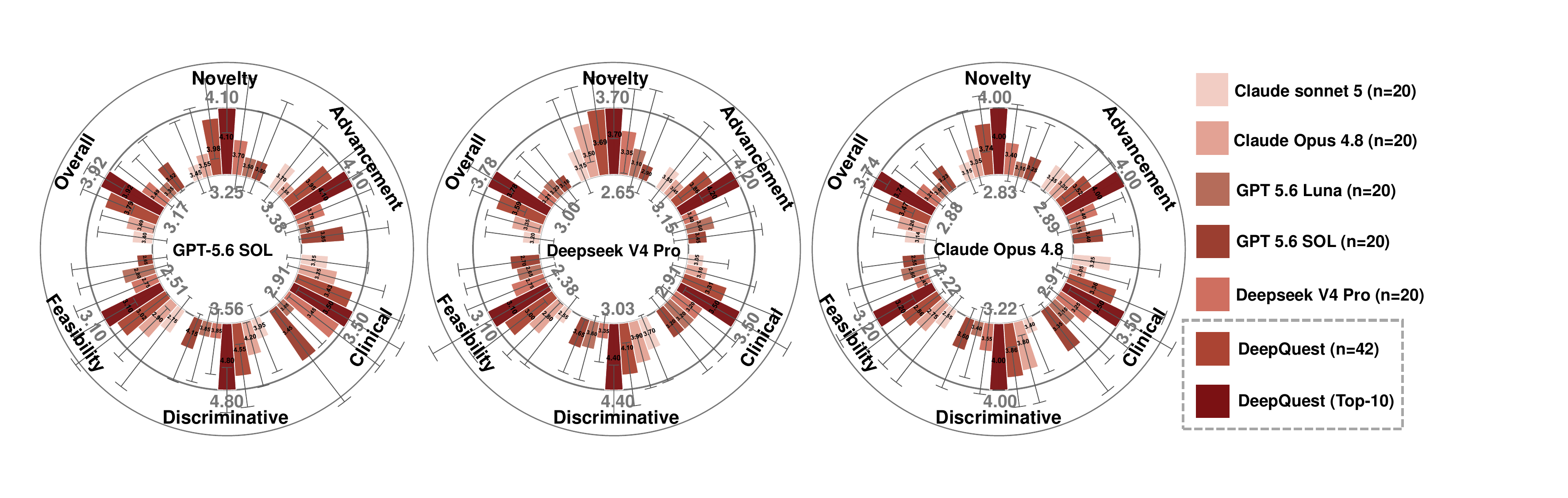}
    \caption{Hypothesis quality evaluation. Bars show mean scores from three independent language model reviewers across multiple evaluation dimensions for baseline methods, DeepQuest, and DeepQuest Top-10, with standard deviations indicated in gray.}
    \label{fig:hypothesis_quality}
\end{figure}

DeepQuest achieves the highest overall mean across all three evaluators, with its strongest gains concentrated in dimensions related to scientific question formulation and empirical discriminative power. Overall improvements over each evaluator’s best baseline are 0.27, 0.24, and 0.21 points, as shown in Fig.~\ref{fig:hypothesis_quality}. DeepQuest outperforms the best baseline in 14 of 15 evaluator–dimension comparisons. The largest gains are 0.35 in empirical discriminative power under GPT-5.6-sol, 0.34 in novelty under Claude Opus 4.8, and 0.26 in scientific advancement under DeepSeek-v4-pro-0813. Clinical significance shows relatively smaller gains than other evaluated dimensions: under GPT-5.6-sol, DeepQuest scores 3.43 versus the best baseline’s 3.45.

Across all three evaluator models, the Top-10 subset scores higher than the full candidate set and exceeds the best corresponding baseline on all five dimensions, indicating alignment between ranking and quality assessments. Because Top-10 is nested within the full set, this provides only descriptive evidence of ranking effectiveness rather than an independent validation of ranking performance.

\FloatBarrier
\subsection{From Hypotheses to Scientific Claims}
\suppressfloats[t]

The full DeepQuest pipeline achieves the highest SFR across the evaluated general-purpose and domain-specific agents. Of 42 tasks, 40 yield Candidate New Discovery or Extension, giving an SFR of 95.24\%, compared with 78.57\% for Codex and 76.19\% for Claude Code on the same DeepQuest-H hypothesis set, as shown in Table~\mbox{\ref{tab:discovery_outcomes_gpt}}. Among the domain-specific baselines, DiscoADRD and Biomni achieve SFRs of 45.45\% and 86.49\%, respectively, on their evaluated task sets. These gains reflect the complete system and cannot be attributed to a single module.

\begin{table}[!ht]
  \centering
  \caption{Discovery outcome comparison across methods. Candidate New denotes Candidate New Discovery. Scientific Finding Rate (SFR) represents the proportion of tasks classified as Candidate New Discovery, Rediscovery, or Extension. Top-10 SFR is computed among the highest-ranked DeepQuest-H tasks. N/A indicates not applicable.}
  \label{tab:discovery_outcomes_gpt}
  \begingroup
  \small
  \color{black}
  \definecolor{DQGroupGray}{HTML}{F2F2F5}
  \definecolor{DQHighlight}{HTML}{EBF0FF}
  \setlength{\tabcolsep}{2pt}
  \setlength{\aboverulesep}{1.5pt}
  \setlength{\belowrulesep}{1.5pt}
  \renewcommand{\arraystretch}{1.12}
  \resizebox{\linewidth}{!}{%
  \begin{tabular}{@{}llccccccc@{}}
    \toprule
    \multirow{2}{*}[-0.5ex]{\textbf{Agent type}} & \multirow{2}{*}[-0.5ex]{\textbf{Method}}
    & \multicolumn{5}{c}{\textbf{Discovery outcome}}
    & \multicolumn{2}{c}{\textbf{Finding rate}} \\
    \cmidrule(lr){3-7}\cmidrule(l){8-9}
    & & \textbf{Candidate New}
    & \textbf{Rediscovery} & \textbf{Extension} & \textbf{Uncertain}
    & \textbf{Skip} & \textbf{SFR} & \textbf{Top-10 SFR} \\
    \midrule

    & \multicolumn{8}{l}{\cellcolor{DQGroupGray}\textbf{GPT-H}\IfFileExists{figures/GPT.pdf}{\,\includegraphics[height=1.7ex]{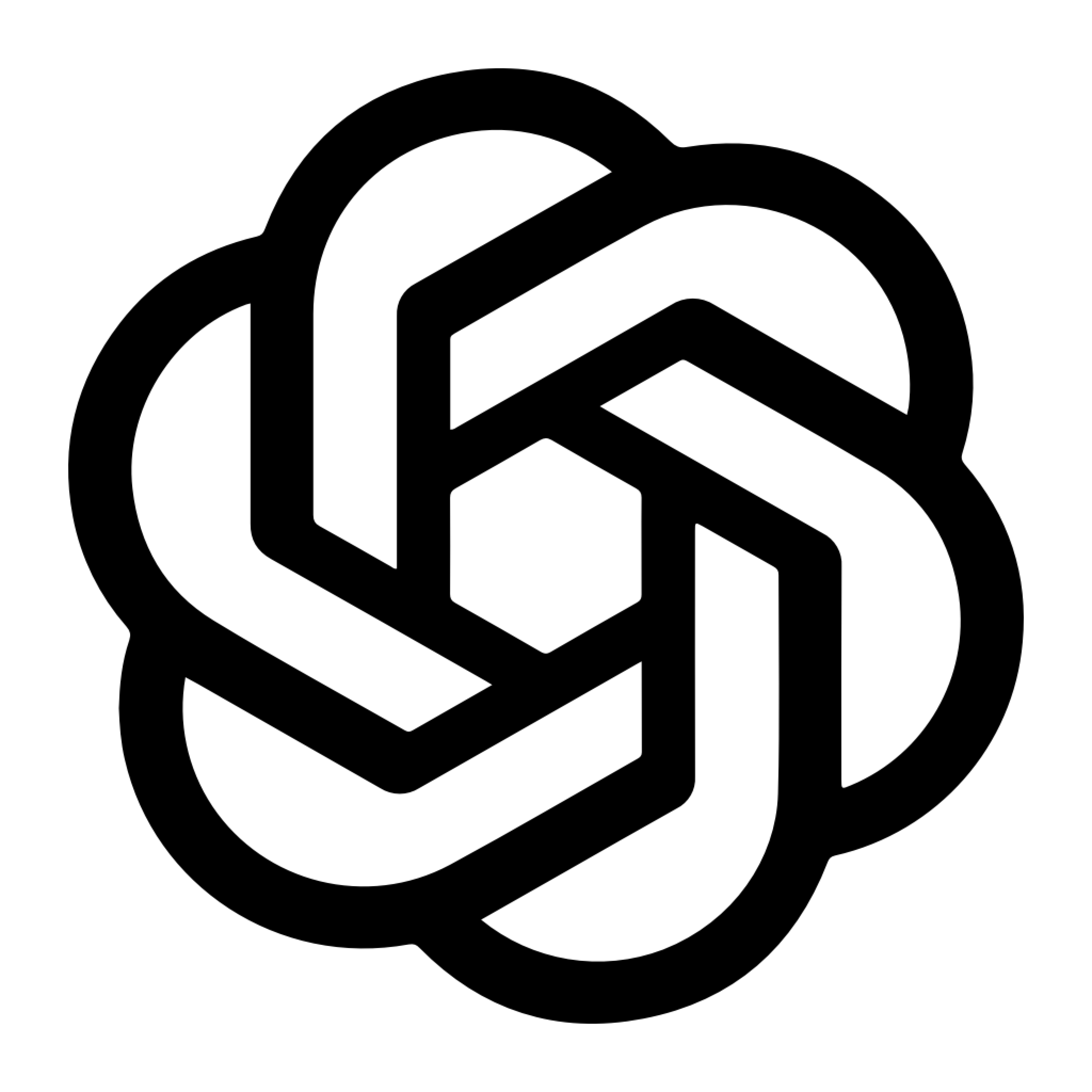}}{}\enspace($n=29$)} \\
    & Codex & 13 (44.83\%) & 5 (17.24\%) & 4 (13.79\%) & 7 (24.14\%) & 0 (0.00\%) & 22 (75.86\%) & N/A \\
    & Claude Code & 7 (24.14\%) & 2 (6.90\%) & 11 (37.93\%) & 2 (6.90\%) & 7 (24.14\%) & 20 (68.97\%) & N/A \\

    \cmidrule(l){2-9}
    & \multicolumn{8}{l}{\cellcolor{DQGroupGray}\textbf{Claude-H}\IfFileExists{figures/claude.pdf}{\,\includegraphics[height=1.7ex]{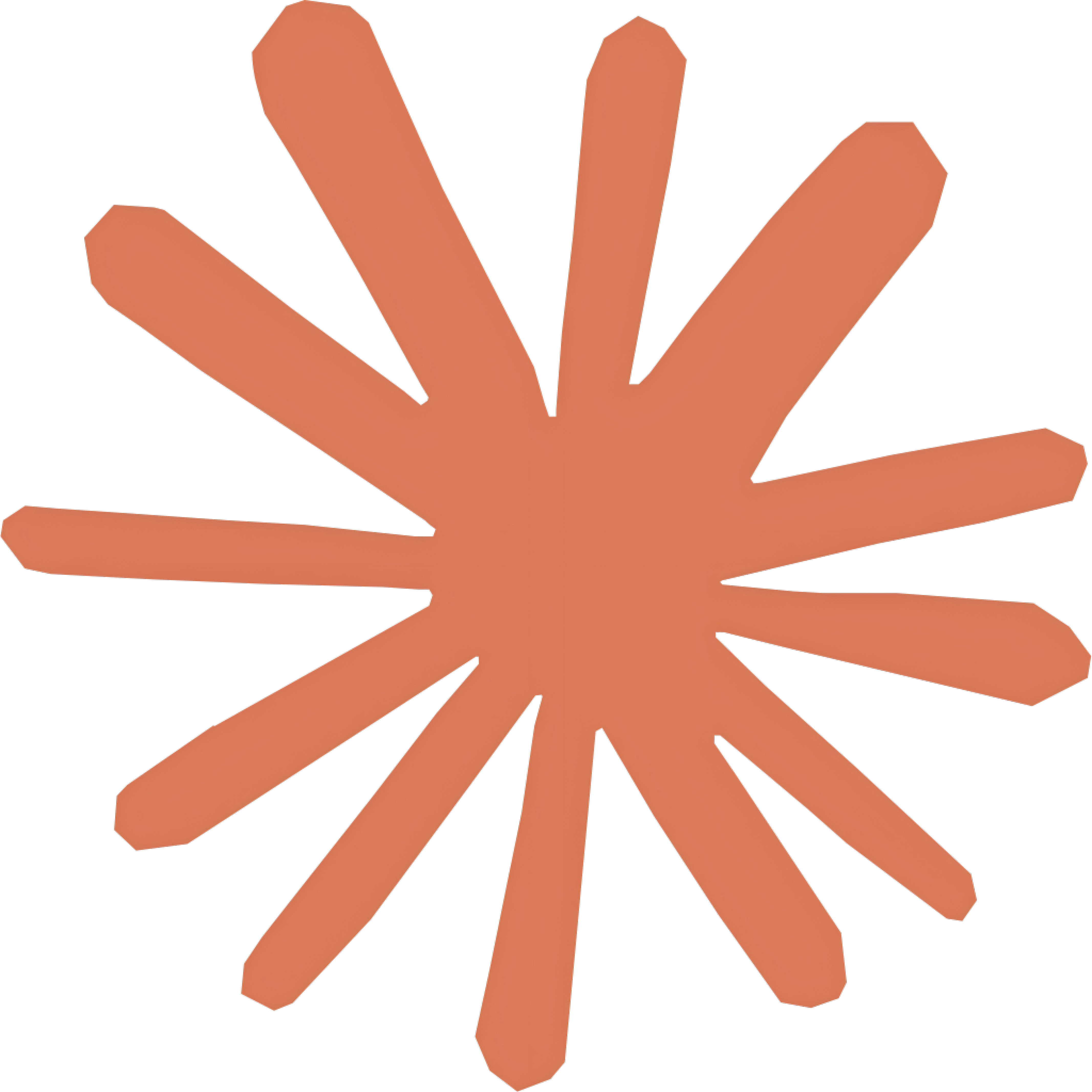}}{}\enspace($n=30$)} \\
    & Codex & 9 (30.00\%) & 0 (0.00\%) & 11 (36.67\%) & 8 (26.67\%) & 2 (6.67\%) & 20 (66.67\%) & N/A \\
    & Claude Code & 10 (33.33\%) & 0 (0.00\%) & 12 (40.00\%) & 2 (6.67\%) & 6 (20.00\%) & 22 (73.33\%) & N/A \\

    \cmidrule(l){2-9}
    & \multicolumn{8}{l}{\cellcolor{DQGroupGray}\textbf{DeepQuest-H}\enspace($n=42$)} \\
    & Codex & 14 (33.33\%) & 2 (4.76\%) & 17 (40.48\%) & 9 (21.43\%) & 0 (0.00\%) & 33 (78.57\%) & 9 (90.00\%) \\
    \multirow{-9}{*}{\shortstack[l]{\textbf{General-}\\\textbf{purpose}}}
    & Claude Code & 20 (47.62\%) & 0 (0.00\%) & 12 (28.57\%) & 2 (4.76\%) & 8 (19.05\%) & 32 (76.19\%) & 8 (80.00\%) \\

    \midrule[\heavyrulewidth]
    & DiscoADRD & 3 (27.27\%) & 1 (9.09\%) & 1 (9.09\%) & 5 (45.45\%) & 1 (9.09\%) & 5 (45.45\%) & N/A \\
    & Biomni & 13 (35.14\%) & 1 (2.70\%) & 18 (48.65\%) & 5 (13.51\%) & 0 (0.00\%) & 32 (86.49\%) & N/A \\

    \rowcolor{DQHighlight}
    \cellcolor{white}\multirow{-3}{*}{\shortstack[l]{\textbf{Domain-}\\\textbf{specific}}}
    & \textbf{DeepQuest} & 20 (47.62\%) & 0 (0.00\%) & 20 (47.62\%) & 1 (2.38\%) & 1 (2.38\%) & 40 (95.24\%) & 10 (100.00\%) \\
    \bottomrule
  \end{tabular}%
  }
  \endgroup

\par

\centering
  \centering
  \caption{Discovery outcome comparison across methods on the expanded literature corpus.}
  \label{tab:discovery_outcomes_gpt_extended}
  \begingroup
  \small
  \color{black}
  \definecolor{DQGroupGray}{HTML}{F2F2F5}
  \definecolor{DQHighlight}{HTML}{EBF0FF}
  \setlength{\tabcolsep}{2pt}
  \setlength{\aboverulesep}{1.5pt}
  \setlength{\belowrulesep}{1.5pt}
  \renewcommand{\arraystretch}{1.12}
  \resizebox{\linewidth}{!}{%
  \begin{tabular}{@{}llccccccc@{}}
    \toprule
    \multirow{2}{*}[-0.5ex]{\textbf{Agent type}} & \multirow{2}{*}[-0.5ex]{\textbf{Method}}
    & \multicolumn{5}{c}{\textbf{Discovery outcome}}
    & \multicolumn{2}{c}{\textbf{Finding rate}} \\
    \cmidrule(lr){3-7}\cmidrule(l){8-9}
    & & \textbf{Candidate New}
    & \textbf{Rediscovery} & \textbf{Extension} & \textbf{Uncertain}
    & \textbf{Skip} & \textbf{SFR} & \textbf{Top-10 SFR} \\
    \midrule

    & \multicolumn{8}{l}{\cellcolor{DQGroupGray}\textbf{GPT-H}\IfFileExists{figures/GPT.pdf}{\,\includegraphics[height=1.7ex]{figures/GPT.pdf}}{}\enspace($n=29$)} \\
    & Codex & 9 (31.03\%) & 8 (27.59\%) & 2 (6.90\%) & 10 (34.48\%) & 0 (0.00\%) & 19 (65.52\%) & N/A \\
    & Claude Code & 2 (6.90\%) & 3 (10.34\%) & 14 (48.28\%) & 3 (10.34\%) & 7 (24.14\%) & 19 (65.52\%) & N/A \\

    \cmidrule(l){2-9}
    & \multicolumn{8}{l}{\cellcolor{DQGroupGray}\textbf{Claude-H}\IfFileExists{figures/claude.pdf}{\,\includegraphics[height=1.7ex]{figures/claude.pdf}}{}\enspace($n=30$)} \\
    & Codex & 6 (20.00\%) & 3 (10.00\%) & 11 (36.67\%) & 8 (26.67\%) & 2 (6.67\%) & 20 (66.67\%) & N/A \\
    & Claude Code & 7 (23.33\%) & 0 (0.00\%) & 12 (40.00\%) & 5 (16.67\%) & 6 (20.00\%) & 19 (63.33\%) & N/A \\

    \cmidrule(l){2-9}
    & \multicolumn{8}{l}{\cellcolor{DQGroupGray}\textbf{DeepQuest-H}\enspace($n=42$)} \\
    & Codex & 10 (23.81\%) & 6 (14.29\%) & 13 (30.95\%) & 13 (30.95\%) & 0 (0.00\%) & 29 (69.05\%) & 9 (90.00\%) \\
    \multirow{-9}{*}{\shortstack[l]{\textbf{General-}\\\textbf{purpose}}}
    & Claude Code & 11 (26.19\%) & 1 (2.38\%) & 16 (38.10\%) & 6 (14.29\%) & 8 (19.05\%) & 28 (66.67\%) & 7 (70.00\%) \\

    \midrule[\heavyrulewidth]
    & DiscoADRD & 2 (18.18\%) & 2 (18.18\%) & 1 (9.09\%) & 5 (45.45\%) & 1 (9.09\%) & 5 (45.45\%) & N/A \\
    & Biomni & 8 (21.62\%) & 1 (2.70\%) & 20 (54.05\%) & 8 (21.62\%) & 0 (0.00\%) & 29 (78.38\%) & N/A \\

    \rowcolor{DQHighlight}
    \cellcolor{white}\multirow{-3}{*}{\shortstack[l]{\textbf{Domain-}\\\textbf{specific}}}
    & \textbf{DeepQuest} & 13 (30.95\%) & 2 (4.76\%) & 25 (59.52\%) & 1 (2.38\%) & 1 (2.38\%) & 40 (95.24\%) & 10 (100.00\%) \\
    \bottomrule
  \end{tabular}%
  }
  \endgroup
\end{table}

Under either independent executor, DeepQuest-H achieves the highest SFR among the three hypothesis sets. Because the sets differ in size and composition, these comparisons reflect combined generation and selection performance rather than an isolated effect of hypothesis quality. Top-10 SFRs were 90\%, 80\%, and 100\% under Codex, Claude Code, and DeepQuest, respectively, exceeding their full-set rates. All ten DeepQuest outcomes were Candidate New Discovery or Extension. Given the small subset, these results provide only descriptive evidence of ranking effectiveness.

DeepQuest maintains its leading performance under the expanded literature corpus, with unchanged SFR and Top-10 SFR as shown in Table~\mbox{\ref{tab:discovery_outcomes_gpt_extended}}. Compared with the literature corpus containing 200 studies, the expanded corpus provides more relevant studies for matching, with the number of DeepQuest outcomes classified as Candidate New Discovery decreasing from 20 to 13, while Extension and Rediscovery increase from 20 to 25 and from 0 to 2. Rediscovery represents evidence-supported replication of existing findings, whereas Extension and Candidate New Discovery indicate potential knowledge gains. The latter two categories still account for 90.48\% of all DeepQuest tasks, demonstrating its continued potential to extend existing knowledge under broader literature coverage.

\FloatBarrier
\subsection{Ablation Study}

As shown in Table~\mbox{\ref{tab:ablation}}, the full DeepQuest framework achieves an SFR of 95.24\% and a Top-10 SFR of 100\%. Removing data feasibility screening leaves 19 tasks incomplete and reduces SFR to 66.67\%, highlighting its importance in ensuring task executability. Disabling execution auditing reduces SFR to 85.71\%, demonstrating its contribution to the reliability of research outcomes. Removing knowledge gap extraction decreases SFR to 92.68\%, indicating the benefit of literature-derived knowledge gaps for scientific discovery. Without Elo ranking, the Top-10 SFR decreases from 100\% to 90\%, suggesting that ranking helps prioritize research tasks with greater discovery potential. Disabling cross-round revision reduces SFR to 90.48\%, confirming the value of iterative refinement. Together, these components improve task feasibility, experimental reliability, and research prioritization, strengthening DeepQuest's overall performance.

\suppressfloats[t]
\par\noindent
\begin{minipage}{\linewidth}
  \centering
  \captionof{table}{Ablation study of DeepQuest. We evaluate the effects of removing data feasibility screening, execution auditing, knowledge gap extraction, Elo ranking, and cross-round revision.}
  \label{tab:ablation}
  \begingroup
  \definecolor{AblationBlue}{HTML}{EBF0FF}
  \small
  \color{black}
  \setlength{\tabcolsep}{2pt}
  \setlength{\aboverulesep}{1.5pt}
  \setlength{\belowrulesep}{1.5pt}
  \renewcommand{\arraystretch}{1.28}
  % Equal columns with less unused width, yielding larger text after scaling.
  \settowidth{\dimen0}{\textbf{Candidate New}}
  \settowidth{\dimen2}{\textbf{Rediscovery}}
  \advance\dimen0 by \dimen2
  \divide\dimen0 by 2
  \settowidth{\dimen2}{10\,(100.00\%)}
  \ifdim\dimen2>\dimen0 \dimen0=\dimen2\fi
  \advance\dimen0 by 0.5pt
  \edef\AblationMetricWidth{\the\dimen0}
  \newcommand{\AblationHead}[1]{\makebox[\AblationMetricWidth][c]{\textbf{#1}}}
  \resizebox{\linewidth}{!}{%
  \begin{tabular}{@{}l*{7}{>{\centering\arraybackslash}p{\AblationMetricWidth}}@{}}
    \toprule
    \multicolumn{1}{@{}l}{\multirow{2}{*}[-3pt]{\fontsize{11}{13}\selectfont\bfseries Method}} & \multicolumn{5}{c}{\textbf{Discovery outcome}}
    & \multicolumn{2}{c}{\textbf{Finding rate}} \\
    \cmidrule(lr){2-6}\cmidrule(l){7-8}
    & \AblationHead{Candidate New} & \AblationHead{Rediscovery}
    & \AblationHead{Extension} & \AblationHead{Uncertain} & \AblationHead{Skip}
    & \AblationHead{SFR} & \AblationHead{Top-10 SFR} \\
    \midrule
    w/o Data Feasibility & 20\,(33.33\%) & 0\,(0.00\%) & 20\,(33.33\%) & 1\,(1.67\%) & 19\,(31.67\%) & 40\,(66.67\%) & 9\,(90.00\%) \\
    w/o Execution Audit & 25\,(59.52\%) & 3\,(7.14\%) & 8\,(19.05\%) & 5\,(11.90\%) & 1\,(2.38\%) & 36\,(85.71\%) & 8\,(80.00\%) \\
    w/o Gap Extraction & 26\,(63.41\%) & 5\,(12.20\%) & 7\,(17.07\%) & 2\,(4.88\%) & 1\,(2.44\%) & 38\,(92.68\%) & 9\,(90.00\%) \\
    w/o Elo Ranking & 20\,(47.62\%) & 0\,(0.00\%) & 20\,(47.62\%) & 1\,(2.38\%) & 1\,(2.38\%) & 40\,(95.24\%) & 9\,(90.00\%) \\
    w/o Cross-round Revision & 23\,(54.76\%) & 4\,(9.52\%) & 11\,(26.19\%) & 3\,(7.14\%) & 1\,(2.38\%) & 38\,(90.48\%) & 10\,(100.00\%) \\
    % Slight color overlap suppresses hairline seams between adjacent cells.
    \rowcolor{AblationBlue}[\dimexpr\tabcolsep+0.2pt\relax][\dimexpr\tabcolsep+0.2pt\relax]
    \textbf{DeepQuest (Ours)} & 20\,(47.62\%) & 0\,(0.00\%) & 20\,(47.62\%) & 1\,(2.38\%) & 1\,(2.38\%) & 40\,(95.24\%) & 10\,(100.00\%) \\
    \bottomrule
  \end{tabular}
  }
  \endgroup
\end{minipage}
\par\addvspace{\baselineskip}

\subsection{Representative Medical Findings and External Replication}

\subsubsection{Neuropsychiatric Symptoms and Subsequent Dementia Risk}

First observed neuropsychiatric symptom positivity was associated with higher incident dementia risk in the two-year lag analysis, as shown in Fig.~\ref{fig:medical_findings}. Using the 10 Neuropsychiatric Inventory Questionnaire (NPI-Q) symptoms, DeepQuest constructed time-varying risk intervals for 0, 365, and 730-day lags and fitted time-varying Cox models to symptom-history states. At the 730-day lag, first observed positivity had a hazard ratio (HR) of 2.496 (95\% CI: 1.528–4.077) in OASIS-3 and 3.326 (1.443–7.668) in ADNI, relative to no prior or current positive symptom record.

In OASIS-3, HR point estimates were 1.847 for prior positivity with current negativity, 2.496 for first observed positivity, and 3.598 for positivity both previously and at the current visit. ADNI did not show the same ordering of point estimates, but first observed symptom positivity was likewise associated with a higher risk of incident dementia under the 730-day lag. 

\subsubsection{Longitudinal Preservation of Relative Regional Amyloid Distribution}

\begin{figure}[ht]
    \centering
    
    \includegraphics[width=\linewidth]{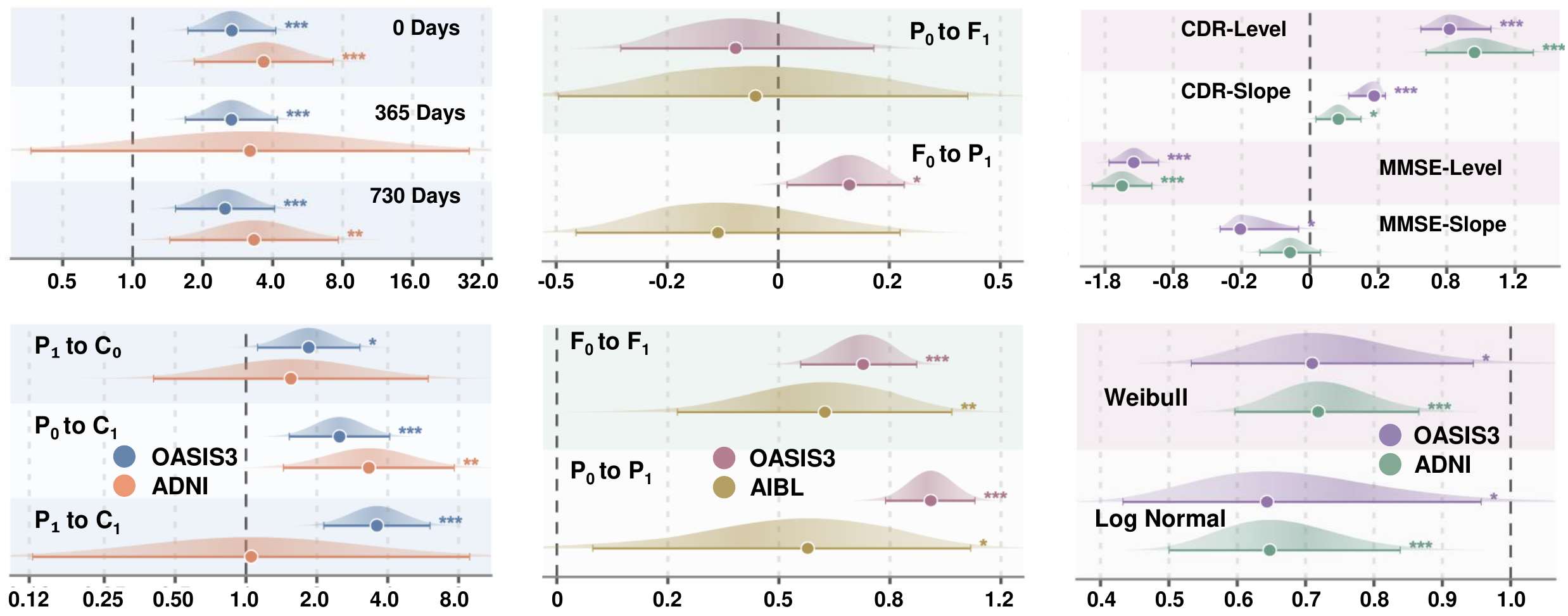}
    \caption{Representative case studies produced by DeepQuest. Results are validated in OASIS-3 with external evaluation in ADNI or AIBL. Left: NPS–dementia associations across lags (upper) and symptom-history states at 730-day lag (lower). Mid: cross- and within-region longitudinal amyloid associations. Right: neocortical tau associations with cognitive levels/slopes (upper) and AFT time ratios for clinical impairment (lower). Dots and bars indicate estimates and 95\% CIs.
    }
    \label{fig:medical_findings}
\end{figure}

DeepQuest investigated whether regional amyloid distribution provides a stable imaging phenotype beyond global amyloid burden, as shown in Fig.~\ref{fig:medical_findings}. Using the first and last eligible PET scans, it applied a centered log-ratio transformation to derive relative amyloid burden measures for the frontal, parietal, and occipital composite regions. Longitudinal preservation was then evaluated after adjusting for global cortical amyloid burden and relevant covariates.

In OASIS-3 (n=284), the within-region mapping coefficients were 0.7276 (95\% CI:0.5587–0.8965) for the frontal region and 0.9471(0.7881–1.1061) for the parietal region. Corresponding estimates in AIBL (n = 159) were 0.6243 (0.2257–1.0229) and 0.5779 (0.0648–1.0910). All four estimates support within-region preservation across visits. The partial R² values for baseline regional configuration were 0.7638 in OASIS-3 and 0.7193 in AIBL. These results support the longitudinal stability of amyloid distribution and its potential as an imaging phenotype beyond global burden.

\subsubsection{Neocortical Tau and Clinical Impairment Trajectories and Timing}

Higher neocortical tau was associated with greater clinical impairment and faster Clinical Dementia Rating Sum of Boxes (CDR-SB) increases, whereas associations with Mini-Mental State Examination (MMSE) change differed across cohorts, as shown in Fig.~\ref{fig:medical_findings}. Using standardized neocortical tau burden from the first tau PET scan, DeepQuest fitted linear mixed-effects models with subject-specific random intercepts and tau-by-time interactions. Both cohorts showed higher CDR-SB and lower MMSE levels with greater neocortical tau. CDR-SB slope associations were 0.1877 (95\% CI:0.1134–0.2619) in OASIS-3 and 0.0831(0.0171–0.1491) in ADNI, supporting an association between higher neocortical tau and faster clinical progression. MMSE slope associations coefficients were -0.2113 (-0.3890, -0.0335) in OASIS-3 and -0.0583 (-0.1470, 0.0304) in ADNI, indicating longitudinal association replication depends on the clinical measure.

Survival models incorporating censoring further linked higher neocortical tau to earlier recorded clinical impairment across the available clinical history. DeepQuest defined the first occurrence of global CDR ($>0$) as the event and fitted accelerated failure time (AFT) models to 405 OASIS-3 and 578 ADNI participants. Weibull time ratios were 0.7097 (95\% CI: 0.5327–0.9454) in OASIS-3 and 0.7185 (0.5964–0.8657) in ADNI. Log-normal models showed effects in the same direction. These results support associations of neocortical tau with clinical progression and earlier impairment. Further implementation details are provided in Appendix ~\ref{app:B}.
%Accelerated failure time (AFT) models incorporating censoring associated higher neocortical tau with earlier impairment across the available clinical history, using cohort-specific shifted time scales (Appendix B). For first global CDR > 0, Weibull time ratios were 0.7097 (95% CI: 0.5327–0.9454; OASIS-3, n = 405) and 0.7185 (0.5964–0.8657; ADNI, n = 578), with concordant directions under log-normal models.

\section{conclusion}

In this work, we introduced DeepQuest, a knowledge-gap-driven AI scientist framework for Alzheimer’s disease research. It integrates hypothesis formulation, data-constrained prioritization, and iterative experimentation with two-level auditing that distinguishes experimental reliability from knowledge contribution. Each candidate contribution is linked to experimental evidence and assessed against two frozen literature corpora, making its evidential basis explicit. Evaluation on OASIS-3 demonstrates improved hypothesis quality and audit-assessed scientific finding rates over the evaluated baselines, with key associations supported in independent ADNI and AIBL cohorts. These results highlight a path beyond automated experimentation: selecting meaningful, answerable questions and assessing what the resulting evidence adds to existing knowledge.

\subsection*{AI use statement}

We used generative AI tools solely for manuscript refinement, including language translation, grammar correction, and improving clarity and readability. AI tools were not used for research ideation, hypothesis generation, experimental design, data analysis, result generation, dataset creation, mathematical derivations, proofs, or scientific claims. All AI-assisted suggestions were reviewed and verified by the authors, who take full responsibility for the final content.

\subsection*{Ethics statement}

Because AI-generated scientific outputs may contain unsupported interpretations or overlook important limitations, DeepQuest is designed with safeguards to promote responsible AI-assisted scientific discovery. All analyses in this work are conducted using publicly available datasets with established data governance and privacy protection procedures. The system does not access or collect additional personally identifiable information.

The findings generated by DeepQuest should be regarded as evidence-supported research hypotheses or candidate scientific insights, rather than final scientific conclusions or clinical recommendations. AI-generated results require independent validation, expert interpretation, and further investigation before being considered for scientific publication or practical application. DeepQuest is intended to augment human scientific reasoning rather than replace expert judgment. The responsibility for experimental design, result interpretation, and final scientific claims remains with human researchers.

\subsection*{Reproducibility statement}

Source code and evaluation artifacts are available at \url{https://anonymous.4open.science/r/DeepQuest-3066/}


\bibliography{main}
\bibliographystyle{iclr2027_conference}

\appendix

\section{Evaluation of Hypothesis Quality and Scientific Discovery}
\label{app:A}

\subsection{Protocol for Hypothesis Quality Evaluation}

\textbf{Experimental Setup.}  DeepQuest included 42 candidate hypotheses that passed data availability screening. For comparison, we selected five language models as baselines: GPT-5.6-luna, GPT-5.6-sol, Claude Sonnet 5, Claude Opus 4.8, and DeepSeek-v4-pro-0813. Each baseline model generated 20 hypotheses. All baseline models received the same 10 Alzheimer’s disease review articles used by DeepQuest as literature input. During question refinement, DeepQuest used automatically generated data summaries to identify research questions supported by the available data. To reduce differences in data context, the official OASIS-3 overview was also provided to all baseline models. The comparison therefore evaluates the respective hypothesis-generation and selection workflows, including differences in data context and feasibility screening.

\textbf{Evaluation Protocol.}  GPT-5.6-sol, Claude Opus 4.8, and DeepSeek-v4-pro-0813 served as independent evaluators. All three models received the same candidate hypotheses and standardized data description, and evaluated each hypothesis across five dimensions: Novelty, Scientific Advancement, Clinical Significance, Empirical Discriminative Power, and Feasibility. Novelty measures the extent to which a hypothesis introduces new insights beyond the provided reviews and known research questions. Scientific Advancement assesses its potential to resolve scientific uncertainty or advance the investigation of unknown relationships. Clinical Significance evaluates its potential value for clinical interpretation, stratification, monitoring, or understanding of outcomes. Empirical Discriminative Power measures whether the available data and analyses can distinguish among alternative explanations or competing expectations. Feasibility assesses whether the required variables, samples, and analytical conditions are attainable within the given data context. Each dimension was rated on a 1 to 5 scale. Overall was calculated as the equally weighted mean of the five dimensions. Results are reported as hypothesis level means and sample standard deviations.

\subsection{Protocol for Scientific Finding Evaluation}

% \subsection{Protocol for Scientific Finding Evaluation}

\textbf{Hypothesis Set Construction.}  Codex and Claude Code were evaluated on hypothesis sets from three sources. DeepQuest-H comprised all 42 DeepQuest hypotheses that passed data availability screening. GPT-H combined hypotheses generated by GPT-5.6-luna and GPT-5.6-sol. Researchers verified the required variables and analytical conditions against the available data assets, retaining 29 hypotheses. Claude-H was constructed analogously by combining hypotheses generated by Claude Opus 4.8 and Claude Sonnet 5, with 30 hypotheses retained. 

\textbf{Baseline Configuration.}  Independent execution environments included Codex powered by GPT-5.6-luna and Claude Code powered by Claude Sonnet 5, both of which were compared with the complete DeepQuest pipeline. Biomni was evaluated as a complete research system, receiving the same review articles. For DiscoADRD, a hypothesis-testing system, GPT-5.6-luna generated hypotheses from the same review articles under the same hypothesis-generation budget.

\textbf{Outcome Evaluation and Top-10 Selection.}  Tables  \mbox{\ref{tab:discovery_outcomes_gpt}},  \mbox{\ref{tab:ablation}}, and \mbox{\ref{tab:discovery_outcomes_claude}} used the frozen corpus of 200 clinical studies, whereas Table  \mbox{\ref{tab:discovery_outcomes_gpt_extended}} used the expanded corpus of 4604 articles. GPT-5.6-luna served as the evaluator for Tables  \mbox{\ref{tab:discovery_outcomes_gpt}}, \mbox{\ref{tab:discovery_outcomes_gpt_extended}} and \mbox{\ref{tab:ablation}}, while Table \mbox{\ref{tab:discovery_outcomes_claude}} used Claude Sonnet 5.0. DeepQuest Top-10 consisted of the highest-priority tasks selected from the 42 DeepQuest-H hypotheses according to groupwise ranking. For the ablation without Elo ranking, Top-10 SFR was computed on 10 hypotheses randomly sampled from the same feasible hypothesis set.

\textbf{Top-10 Outcomes.}  For the DeepQuest Top-10 subset under the evaluation setting of Table~\mbox{\ref{tab:discovery_outcomes_gpt}}, Codex produced auditable conclusions for 9 of 10 tasks, yielding an SFR of 90\%. Across all 10 tasks, 5 were classified as Candidate New Discovery, 3 as Extension, 1 as Rediscovery, and 1 as Uncertain. Claude Code produced auditable conclusions for 8 of 10 tasks, yielding an SFR of 80\%. Its 10 task outcomes comprised 4 Candidate New Discovery, 4 Extension, 1 Uncertain, and 1 Skip. DeepQuest produced auditable conclusions for all 10 tasks, yielding an SFR of 100\%. Of these, 4 were classified as Candidate New Discovery and 6 as Extension, with no Rediscovery, Uncertain, or Skip outcomes.

%\subsection{Preference Analysis}

\subsection{Preference Analysis}

\textbf{Hypothesis Judge.}  As shown in Fig.~\mbox{\ref{fig:hypothesis_quality}}, DeepQuest consistently ranked highest across evaluators from different model families, without showing a pronounced advantage attributable to same-family evaluation. Specifically, GPT-5.6-sol assigned its own hypotheses an overall score of 3.52, slightly higher than the scores assigned to other baselines (3.35–3.49). Similarly, Claude Opus 4.8 scored its own hypotheses at 3.26, compared with 3.08–3.23 for other baselines. In contrast, DeepSeek-v4-pro-0813 assigned its own hypotheses a score of 3.21, lower than its score for hypotheses generated by Claude Opus 4.8 (3.35), indicating that self-preference was not consistently observed across evaluators. Although DeepQuest-H was generated using a GPT-family model, its margin over the strongest baseline was 0.27 under the GPT judge and remained 0.21 and 0.24 under the Claude and DeepSeek judges, respectively, showing no pronounced same-family scoring advantage. These results suggest that, despite some evaluator-specific self-preference tendencies, DeepQuest maintained a consistent relative advantage across model families.

\textbf{Scientific Finding Evaluation.}  As shown in Tables~\mbox{\ref{tab:discovery_outcomes_gpt}} and~\mbox{\ref{tab:discovery_outcomes_claude}}, DeepQuest consistently achieves the highest SFR (95.24\%) and Top-10 SFR (100\%) under both GPT 5.6 luna and Claude Sonnet 5.0 evaluation, maintaining its relative advantage across evaluator model families. Under Claude evaluation, the SFR increases for all six general-purpose agent configurations, with gains of 3.45–10.34 and 3.33–16.67 percentage points for GPT-H and Claude-H, respectively, indicating that the improvements are not limited to hypotheses generated by models from the same family. Regarding contribution labels, Claude shows a greater tendency than GPT to classify outcomes as Candidate New Discovery, with increases in this category across all methods and configurations except DiscoADRD. Both Candidate New Discovery and Extension represent potential knowledge gains relative to the reference evidence. Both evaluators assign 40 DeepQuest outcomes to these two categories, further supporting its consistent advantage in scientific finding rate and knowledge contribution.

\begin{table}[t]
  \centering
  \caption{Discovery outcome comparison across methods under Claude evaluation.}
  \label{tab:discovery_outcomes_claude}
  \begingroup
  \small
  \color{black}
  \definecolor{DQGroupGray}{HTML}{F2F2F5}
  \definecolor{DQHighlight}{HTML}{EBF0FF}
  \setlength{\tabcolsep}{2pt}
  \setlength{\aboverulesep}{1.5pt}
  \setlength{\belowrulesep}{1.5pt}
  \renewcommand{\arraystretch}{1.12}
  \resizebox{\linewidth}{!}{%
  \begin{tabular}{@{}llccccccc@{}}
    \toprule
    \multirow{2}{*}[-0.5ex]{\textbf{Agent type}} & \multirow{2}{*}[-0.5ex]{\textbf{Method}}
    & \multicolumn{5}{c}{\textbf{Discovery outcome}}
    & \multicolumn{2}{c}{\textbf{Finding rate}} \\
    \cmidrule(lr){3-7}\cmidrule(l){8-9}
    & & \textbf{Candidate New}
    & \textbf{Rediscovery} & \textbf{Extension} & \textbf{Uncertain}
    & \textbf{Skip} & \textbf{SFR} & \textbf{Top-10 SFR} \\
    \midrule
    & \multicolumn{8}{l}{\cellcolor{DQGroupGray}\textbf{GPT-H}\IfFileExists{figures/GPT.pdf}{\,\includegraphics[height=1.7ex]{figures/GPT.pdf}}{}\enspace($n=29$)} \\
    & Codex & 21 (72.41\%) & 4 (13.79\%) & 0 (0.00\%) & 4 (13.79\%) & 0 (0.00\%) & 25 (86.21\%) & N/A \\
    & Claude Code & 13 (44.83\%) & 3 (10.34\%) & 5 (17.24\%) & 1 (3.45\%) & 7 (24.14\%) & 21 (72.41\%) & N/A \\
    \cmidrule(l){2-9}
    & \multicolumn{8}{l}{\cellcolor{DQGroupGray}\textbf{Claude-H}\IfFileExists{figures/claude.pdf}{\,\includegraphics[height=1.7ex]{figures/claude.pdf}}{}\enspace($n=30$)} \\
    & Codex & 15 (50.00\%) & 2 (6.67\%) & 8 (26.67\%) & 3 (10.00\%) & 2 (6.67\%) & 25 (83.33\%) & N/A \\
    & Claude Code & 19 (63.33\%) & 1 (3.33\%) & 3 (10.00\%) & 1 (3.33\%) & 6 (20.00\%) & 23 (76.67\%) & N/A \\
    \cmidrule(l){2-9}
    & \multicolumn{8}{l}{\cellcolor{DQGroupGray}\textbf{DeepQuest-H}\enspace($n=42$)} \\
    & Codex & 25 (59.52\%) & 4 (9.52\%) & 9 (21.43\%) & 4 (9.52\%) & 0 (0.00\%) & 38 (90.48\%) & 9 (90.00\%) \\
    \multirow{-9}{*}{\shortstack[l]{\textbf{General-}\\\textbf{purpose}}} & Claude Code & 31 (73.81\%) & 1 (2.38\%) & 1 (2.38\%) & 1 (2.38\%) & 8 (19.05\%) & 33 (78.57\%) & 9 (90.00\%) \\
    \midrule[\heavyrulewidth]
    & DiscoADRD  & 3 (27.27\%) & 1 (9.09\%) & 1 (9.09\%) & 5 (45.45\%) & 1 (9.09\%) & 5 (45.45\%) & N/A \\
    & Biomni  & 24 (64.86\%) & 4 (10.81\%) & 3 (8.11\%) & 6 (16.22\%) & 0 (0.00\%) & 31 (83.78\%) & N/A \\
    \rowcolor{DQHighlight}
    \cellcolor{white}\multirow{-3}{*}{\shortstack[l]{\textbf{Domain-}\\\textbf{specific}}} & \textbf{DeepQuest} & 33 (78.57\%) & 0 (0.00\%) & 7 (16.67\%) & 1 (2.38\%) & 1 (2.38\%) & 40 (95.24\%) & 10 (100.00\%) \\
    \bottomrule
  \end{tabular}%
  }
  \endgroup
\end{table}

\section{Additional Details on Representative Medical Findings}
\label{app:B}

\subsection{Neuropsychiatric Symptoms and Subsequent Dementia Risk}

\subsubsection{Symptom Definition and Sample Inclusion}

Neuropsychiatric symptoms were defined using the ten symptom items of the NPI-Q. A visit was used to update exposure status only when all ten items had valid records, the visit was marked as a complete NPI-Q assessment, and the participant had not developed dementia by that visit. NPI-Q records obtained after dementia onset were not used to define exposure. Based on current and prior records, symptom status was classified into four categories: negative both previously and currently, previously positive but currently negative, previously negative but currently positive, and positive both previously and currently.

The OASIS-3 input data comprised 7,172 records from 1,071 participants. Among them, 1,052 participants contributed 5,947 candidate visits that met the criteria for dementia-free status and complete NPI-Q assessment. After applying lags of 0, 365, and 730 days, the three risk sets included 880 participants with 190 events, 868 participants with 178 events, and 837 participants with 149 events, respectively. The raw ADNI data comprised 4,982 records from 593 participants. Among them, 425 participants had complete NPI-Q assessments, and 117 incident dementia events were identified.

\subsubsection{Risk Intervals and Statistical Models}

Lag windows of 0, 365, and 730 days were evaluated separately. For each window, counting-process risk intervals were reconstructed. Each risk interval began on the date of the current complete NPI-Q visit plus the corresponding lag and ended at the earliest of the next complete NPI-Q visit plus that lag, the participant’s last observation date, or the first dementia date. An event was recorded when the interval ended at the first dementia date. Only intervals beginning before dementia onset contributed risk time. Consequently, the three lag windows yielded distinct risk sets rather than simple temporal shifts of exposure dates within an identical sample.

The OASIS-3 analyses at lags of 0, 365, and 730 days included 880 participants with 190 events, 868 participants with 178 events, and 837 participants with 149 events, respectively. The corresponding ADNI analyses included 256 participants with 35 events, 226 participants with 33 events, and 200 participants with 29 events.

Symptom history was analyzed using a time-varying Cox proportional hazards model, with participants who were negative both previously and currently as the reference group. Models adjusted for baseline age, sex, years of education, and cumulative vascular status, with participant-clustered robust standard errors. Cumulative vascular status was derived from medical history recorded up to the corresponding NPI-Q exposure visit. Intervals with unknown vascular status were excluded.

\subsection{Longitudinal Preservation of Relative Regional Amyloid Distribution}

\subsubsection{Regional Measures and Inclusion of Repeated PET Scans}

The analysis summarized amyloid burden across the frontal, parietal, and prespecified occipital composite regions. The occipital composite comprised the cuneus, lateral occipital cortex, and lingual gyrus. For each PET visit, the mean standardized uptake value ratio (SUVR) was first calculated for each of the three composite regions, followed by a centered log-ratio transformation:
\begin{equation}
\operatorname{clr}(r_k)
=
\log(r_k)
-
\frac{1}{3}
\sum_{j=1}^{3}
\log(r_j),
\label{eq:clr}
\end{equation}
where $r_k$ denotes the SUVR of the $k$-th composite region, and $clr(r_k)$ denotes its centered log-ratio transformed value. Because the three $clr$ coordinates sum to zero by construction, only the frontal and parietal coordinates were retained as two nonredundant dimensions.

OASIS-3 included participants with at least two eligible amyloid PET scans and complete regional and global cortical SUVR measurements. The first and last eligible PET scans were paired, yielding a final sample of 284 participants. For AIBL, the two PET scans were required to come from the same participant, use the same tracer, and fall within the prespecified intervisit interval. The PiB-only analysis ultimately included 159 participants.

\subsubsection{Multivariate Topographic Model}

The follow-up frontal and parietal   coordinates were modeled jointly as outcomes, with the baseline frontal and parietal   coordinates as the primary predictors. The model additionally adjusted for sex, scan interval, and global cortical amyloid burden at both baseline and follow-up. Tracer type was further adjusted for when multiple tracers were present. The model can be written as:

\begin{equation}
\begin{pmatrix}
F_1 \\
P_1
\end{pmatrix}
=
\mathbf{B}
\begin{pmatrix}
F_0 \\
P_0
\end{pmatrix}
+
\mathbf{\Gamma} C
+
\boldsymbol{\varepsilon},
\label{eq:joint_topology_model}
\end{equation}

where $F_0$ and $P_0$ denote the baseline frontal and parietal clr-transformed coordinates,
$F_1$ and $P_1$ denote the corresponding follow-up coordinates, and $C$ denotes
the covariates.

In OASIS-3, the cross-regional coefficient for $F_1 \leftarrow P_0$ was
$-0.0767$ ($95\%$ CI: $-0.3256$--$0.1722$). The coefficient for
$P_1 \leftarrow F_0$ was $0.1282$ ($0.0164$--$0.2404$).
The joint test of the four baseline topographic coefficients yielded
$\chi^2(4)=509.94$, $P=4.74\times10^{-109}$.

In AIBL, the coefficient for $F_1 \leftarrow P_0$ was
$-0.0409$ ($95\%$ CI: $-0.4938$--$0.4121$). The coefficient for
$P_1 \leftarrow F_0$ was $-0.1086$ ($-0.4465$--$0.2292$).
The joint test yielded $\chi^2(4)=51.440$, $P=1.81\times10^{-10}$.

\subsection{Neocortical Tau and Clinical Impairment Trajectories and Timing}

\subsubsection{Linear Mixed-Effects Models of Cognitive Trajectories}

Across both cohorts, time was defined in years relative to the first eligible tau PET scan (date difference divided by 365.25), and only post-scan cognitive assessments were included. CDR-SB and MMSE were modeled in their original score units.

In OASIS-3, the first eligible tau PET scan for each participant was defined as baseline and matched to the nearest quantitative amyloid PET scan within 365 days. Medial temporal and neocortical tau were defined as mean SUVR across the entorhinal cortex and hippocampus, and across the inferior parietal cortex, precuneus, and temporal pole, respectively. Tau measures were standardized across participants with imaging data, whereas amyloid burden was standardized within tracer. The complete-case CDR-SB and MMSE models included 981 observations from 374 participants and 577 observations from 301 participants, respectively, and were fitted using linear mixed-effects models with participant-specific random intercepts.

In ADNI, the earliest tau PET scan that passed quality control was used for each participant. The CDR-SB model included 1,697 observations from 561 participants, and the MMSE model included 1,564 observations from 558 participants. Models adjusted for medial temporal tau, age, sex, and years of education. Because a matched quantitative amyloid measure was unavailable, amyloid burden, APOE status, imaging-pair interval, and amyloid tracer were not included as covariates. Accordingly, ADNI served as an external replication under a reduced covariate specification and should not be described as using a model identical to that of OASIS-3.

\subsubsection{Interval-Censored Accelerated Failure Time Model}

Event-time bounds were measured in days relative to the first eligible tau PET scan. Before fitting, all finite bounds were shifted by a cohort-specific constant (5,609 days in OASIS-3 and 5,231 days in ADNI) to ensure positivity. Reported time ratios therefore refer to the shifted time scale.

The first occurrence of global CDR greater than 0 was defined as the clinical impairment event. If a CDR of 0 had been recorded before the first positive assessment, the event time was considered to lie between the last normal assessment and the first positive assessment and was treated as interval-censored. Participants whose first valid assessment was already positive were treated as left-censored. Participants who remained normal throughout follow-up were treated as right-censored at their last normal assessment.

OASIS-3 included 405 participants, comprising 57 left-censored, 41 interval-censored, and 307 right-censored cases. The base model included standardized amyloid burden, medial temporal tau, the date difference between the paired amyloid and tau PET scans, and amyloid tracer. The full model additionally included neocortical tau:
\begin{equation}
M_0:\quad
\operatorname{Surv}(L,R)
\sim
\text{amyloid}
+
\text{MTL Tau}
+
\text{pairing interval}
+
\text{tracer},
\label{eq:aft_reduced}
\end{equation}
\begin{equation}
M_1:\quad
M_0
+
\text{neoTau},
\label{eq:aft_full}
\end{equation}

where $L$ and $R$ denote the lower and upper bounds of the impairment event time. $M_0$ is the reduced model, and $M_1$ is the full model including neocortical tau. ADNI included 578 participants, comprising 228 left-censored, 106 interval-censored, and 244 right-censored cases. The full model included neocortical tau, medial temporal tau, age, sex, and years of education. The reduced model excluded neocortical tau.

Both cohorts were modeled using Weibull and log-normal distributions, with the distribution selected within each cohort according to the AIC of the full model. In OASIS-3, the AIC values were 629.95 for the Weibull model and 646.11 for the log-normal model. In ADNI, the corresponding values were 1463.64 and 1497.68. The Weibull model yielded the lower AIC in both cohorts.

% \subsection{Knowledge Contributions in the Context of Prior Research}

\subsection{Knowledge Contributions in the Context of Prior Research}

To further examine DeepQuest's knowledge contributions, we compared three representative cases with relevant earlier studies, focusing on the specific scientific relationships examined beyond prior findings. We also considered the existing evidence to contextualize the discovery labels assigned by the automated knowledge audit. Interestingly, the discovery labels for all three cases remained consistent across the two frozen literature corpora.

Case 1: Neuropsychiatric Symptoms and Subsequent Dementia Risk. \mbox{\citet{guan2023persistence}} demonstrated that persistent NPS are associated with an increased risk of dementia, while \mbox{\citet{ismail2023optimizing}} further established associations between mild behavioral impairment (MBI), AD biomarkers, and dementia progression. However, prior studies have primarily focused on symptom persistence and overall risk, without fully characterizing risk differences across distinct symptom histories within specific temporal windows. Building on these findings, DeepQuest distinguished between past and current symptom states and introduced different time-lag windows to investigate the dynamic association between symptom changes and subsequent dementia risk. Results from both the OASIS-3 and ADNI cohorts supported a lagged association between newly observed symptom positivity and subsequent dementia risk. This finding extends the established NPS–dementia risk relationship to differences associated with symptom history and temporal lag, consistent with the Extension label assigned by the automated knowledge audit.

Case 2: Longitudinal Preservation of Relative Regional Amyloid Distribution. \mbox{\citet{amemiya2026spatiotemporal}} demonstrated the stability of distinct spatial amyloid deposition subtypes and their associations with cognitive decline and rates of amyloid accumulation. DeepQuest further investigated whether continuous relative regional amyloid distributions show longitudinal preservation after accounting for global amyloid burden. Using a centered log-ratio transformation, DeepQuest found evidence of this preservation in both OASIS-3 and AIBL after adjustment for baseline and follow-up global amyloid burden. The studies considered here did not directly test this specific proposition, helping explain the Candidate New Discovery label assigned by the automated knowledge audit.

Case 3: Neocortical Tau and Time to Clinical Impairment. \mbox{\citet{chen2021staging}} and \mbox{\citet{ossenkoppele2022amyloid}} linked tau pathological stages or positivity to cognitive decline and clinical progression. However, the relationship between continuous neocortical tau burden and the timing of clinical impairment remains less well characterized. DeepQuest examined continuous neocortical tau burden in relation to impairment severity, longitudinal trajectories, and time to first documented clinical impairment. Using accelerated failure time models accounting for censoring, it found that higher tau burden was associated with a shorter time to first documented impairment in both OASIS-3 and ADNI. These results extend prior stage- and positivity-based findings to the relationship between continuous pathological burden and the timing of clinical impairment, consistent with the Extension label assigned by the automated knowledge audit.

\section{Finding Reproduction}

\subsection{Experimental Setup}

We use published studies as external references and evaluate DeepQuest through finding reproduction, focusing on its understanding of scientific questions, experimental reasoning, and analytical execution. We select \citet{shimizu2026florzolotau}, which studied spatial associations between regional tau deposition and cognition, and \citet{heikal2026outcome}, which examined differences between clinical diagnosis and continuous cognitive outcomes. The goal is to test whether DeepQuest can independently recover key scientific relationships under available data constraints.

We use GPT-5.6 to abstract each study into a high level Limitation–Hypothesis (L–H) pair and provide an initial experiment design. The L–H pair preserves the original knowledge gap and target relationship, while the initial design specifies only the analysis objective and comparison direction. DeepQuest then refines the experiment autonomously. Because it receives an abstract scientific question rather than the original analysis pipeline, finding reproduction evaluates whether the system can recover the key scientific relationship through an independently designed experiment, rather than replicate the original procedure step by step. DeepQuest subsequently operationalizes the hypothesis, designs and executes the analysis using OASIS-3, and compares the resulting findings with the core conclusions of the original study.

\subsection{Regional Tau Deposition and Cognitive Function}

\citet{shimizu2026florzolotau} used Florzolotau (18F) PET to examine spatial associations between regional tau deposition and neuropsychological performance across the AD continuum. Higher tau retention was generally associated with poorer cognition, with prominent signals in posterior cortical regions including the temporal and parietal lobes. The study performed voxelwise analyses in SPM12, normalized tau PET to Montreal Neurological Institute (MNI) space, and computed standardized uptake value ratios (SUVRs). Voxelwise regressions adjusted for age and sex were combined with cluster level family wise error (FWE) correction. Memory measures showed the largest tau associated clusters, with 28,425 voxels for Logical Memory delayed recall (LM-delay) and 28,210 for Logical Memory immediate recall (LM-immediate). Attention and executive measures yielded 21,149 voxels for Trail Making Test Part B (TMT-B) and 5,662 for Part A (TMT-A). In item level analyses, ADAS-Cog word recall and word recognition were mainly associated with left temporal tau retention. MMSE registration and delayed recall were primarily associated with bilateral temporal tau, whereas calculation was mainly associated with the left temporal region.

DeepQuest operationalized this question in OASIS-3 as an association analysis between regional tau burden and cognitive performance. Using AV-1451 PET, the system selected five regions of interest (ROIs): entorhinal cortex, hippocampus, temporal pole, inferior parietal cortex, and precuneus. Memory was assessed with Selective Reminding Test–Free Recall (SRTFree), while attention and executive function were assessed with TMT-A. Models adjusted for age, sex, and education, used HC3 robust standard errors, and applied Benjamini–Hochberg correction across all primary region-by-measure comparisons.

\begin{table}[t]
  \centering
  \caption{Regional tau associations with cognitive performance. $\beta$ denotes regression coefficients with 95\% confidence intervals, $p$ and $q$ denote unadjusted and FDR-adjusted p-values, respectively. BH correction was applied across the full family of region-by-measure comparisons, including comparisons not displayed in this table.}
  \label{tab:neuropsychological-brain-associations}

  \begingroup
  \small
  \color{black}
  \setlength{\tabcolsep}{2pt}
  \setlength{\aboverulesep}{2.5pt}
  \setlength{\belowrulesep}{2.5pt}
  \renewcommand{\arraystretch}{1.15}

  % Typography: 10pt headers and 9pt bold region names.
  \newcommand{\tablehead}[1]{{\normalsize\bfseries #1}}
  \newcommand{\regionname}[1]{{\small\bfseries #1}}

  % Align coefficients and confidence interval endpoints.
  \newcommand{\coefci}[3]{%
    \mbox{%
      \makebox[3.5em][r]{$#1$}\,
      $[$\makebox[3.5em][r]{$#2$}$,\,$%
      \makebox[3.5em][r]{$#3$}$]$%
    }%
  }

  % Compact scientific notation with aligned multiplication signs.
  \newcommand{\pqvalue}[2]{%
    \mbox{%
      \makebox[2.5em][r]{$#1$}%
      \(\,\times\,\)%
      \makebox[2.2em][l]{$10^{#2}$}%
    }%
  }

  \begin{tabular*}{\linewidth}
    {@{\extracolsep{\fill}}llccc@{}}
    \toprule
    \tablehead{Brain region}
    & \tablehead{Test}
    & {\normalsize $\bm{\beta}$ \textbf{[95\% CI]}}
    & \tablehead{\textit{p}}
    & \tablehead{\textit{q}} \\
    \midrule

    \multirow{2}{*}{\regionname{Entorhinal cortex}}
    & {\normalsize SRTFree}
    & \coefci{-0.3485}{-0.4525}{-0.2445}
    & \pqvalue{5.14}{-11}
    & \pqvalue{1.03}{-9} \\

    & {\normalsize TMT-A}
    & \coefci{+0.1656}{0.0124}{0.3188}
    & \pqvalue{3.415}{-2}
    & \pqvalue{7.588}{-2} \\
    \midrule[0.25pt]

    \multirow{2}{*}{\regionname{Hippocampus}}
    & {\normalsize SRTFree}
    & \coefci{-0.1973}{-0.3187}{-0.0760}
    & \pqvalue{1.435}{-3}
    & \pqvalue{4.100}{-3} \\

    & {\normalsize TMT-A}
    & \coefci{+0.0768}{-0.0204}{0.1741}
    & \pqvalue{1.213}{-1}
    & \pqvalue{2.425}{-1} \\
    \midrule[0.25pt]

    \multirow{2}{*}{\regionname{Temporal pole}}
    & {\normalsize SRTFree}
    & \coefci{-0.2968}{-0.4010}{-0.1925}
    & \pqvalue{2.41}{-8}
    & \pqvalue{2.41}{-7} \\

    & {\normalsize TMT-A}
    & \coefci{+0.1610}{0.0204}{0.3016}
    & \pqvalue{2.485}{-2}
    & \pqvalue{6.212}{-2} \\
    \midrule[0.25pt]

    \multirow{2}{*}{\regionname{Inferior parietal}}
    & {\normalsize SRTFree}
    & \coefci{-0.2945}{-0.4573}{-0.1316}
    & \pqvalue{3.93}{-4}
    & \pqvalue{1.573}{-3} \\

    & {\normalsize TMT-A}
    & \coefci{+0.4611}{0.2115}{0.7107}
    & \pqvalue{2.93}{-4}
    & \pqvalue{1.467}{-3} \\
    \midrule[0.25pt]

    \multirow{2}{*}{\regionname{Precuneus}}
    & {\normalsize SRTFree}
    & \coefci{-0.2590}{-0.3806}{-0.1375}
    & \pqvalue{2.96}{-5}
    & \pqvalue{1.97}{-4} \\

    & {\normalsize TMT-A}
    & \coefci{+0.4774}{0.2026}{0.7522}
    & \pqvalue{6.61}{-4}
    & \pqvalue{2.205}{-3} \\

    \bottomrule
  \end{tabular*}
  \endgroup
\end{table}

The memory results broadly reproduced the spatial pattern reported in the original study. Higher tau burden in all five regions was associated with poorer free recall and remained significant after FDR correction. The strongest association was observed in the entorhinal cortex, with $\beta=-0.3485$, followed by the temporal pole, inferior parietal cortex, precuneus, and hippocampus, with $\beta=-0.2968$, -0.2945, -0.2590, and -0.1973, respectively. These results support the reported association between temporal tau and poorer memory and suggest that the relationship may extend beyond the medial temporal lobe. Attention and executive function showed a more posterior cortical pattern. TMT-A was most strongly associated with the inferior parietal cortex, with $\beta=0.4611$, and precuneus, with $\beta=0.4774$. Associations in the entorhinal cortex, temporal pole, and hippocampus were not significant after FDR correction. This spatial pattern is consistent with the posterior cortical distribution of TMT related tau associations reported in the original study.

DeepQuest reproduced the core finding that higher regional tau burden is associated with poorer cognitive performance. Across different data and analytical settings, it preserved the spatial pattern of stronger temporal associations for memory and stronger posterior association cortex involvement for attention and executive function. The original study used Florzolotau (18F) PET and voxelwise spatial analysis, whereas DeepQuest evaluated the corresponding relationship using AV-1451 PET, predefined ROIs, and cognitive measures available in OASIS-3 that approximate the original assessments. Therefore, the results are better interpreted as cross-setting validation of the reported finding across different datasets, tau tracers, and analytical settings, rather than as an exact reproduction of the original experimental pipeline.

\subsection{Clinical Impairment Status and Outcome Specificity of Continuous Cognitive Performance}

\citet{heikal2026outcome} showed that clinical dementia diagnosis and continuous cognitive performance are related but capture distinct dimensions of disease phenotype. The same factor may influence the risk of crossing a clinical diagnostic threshold, while others may affect only continuous cognitive performance. Using 660 participants with mild cognitive impairment (MCI) or dementia from the Egyptian Dementia Network, the study applied logistic regression to dementia diagnosis and linear regression to continuous MMSE scores. Education and social engagement were associated with both outcomes, whereas several vascular, metabolic, and genetic factors were associated only with MMSE performance. These findings indicate distinct association patterns for binary clinical status and continuous cognitive function.

\begin{table}[ht]
  \centering
  \caption{Outcome-specific association patterns for clinical and cognitive measures. OR and $\beta$ denote odds ratios and regression coefficients for the respective outcomes with 95\% confidence intervals. Age and education are expressed per standard deviation. Comparisons are APOE $\varepsilon4$: $\varepsilon4$/$\varepsilon4$ vs $\varepsilon3$/$\varepsilon3$. Race: Black vs White. SES denotes socioeconomic status. ANXIETY and OCD denote a diagnosis or history of anxiety and obsessive-compulsive disorder, respectively. ELAT denotes elation or elevated mood. FOCLDEF denotes focal neurological deficits. $q$ denotes the BH-adjusted $p$-value across all tested comparisons within each factor group and outcome.}
  \label{tab:cdr-mmse-associations}
  \par\vspace{4pt}

  \begingroup
  \small
  \setlength{\tabcolsep}{2.5pt}
  \setlength{\aboverulesep}{3pt}
  \setlength{\belowrulesep}{3pt}
  \renewcommand{\arraystretch}{1.30}

  % Align OR estimates and confidence interval endpoints.
  \newcommand{\orci}[3]{%
    \mbox{%
      \makebox[2.6em][r]{$#1$}\,
      $[$\makebox[2.6em][r]{$#2$}$,\,$%
      \makebox[3.1em][r]{$#3$}$]$%
    }%
  }

  % Align beta estimates and confidence interval endpoints.
  \newcommand{\betaci}[3]{%
    \mbox{%
      \makebox[3.2em][r]{$#1$}\,
      $[$\makebox[3.2em][r]{$#2$}$,\,$%
      \makebox[3.2em][r]{$#3$}$]$%
    }%
  }

  % Align mantissas and multiplication signs.
  \newcommand{\sci}[2]{%
    \mbox{%
      \makebox[1.8em][l]{$#1$}%
      \(\,\times\,\)%
      \makebox[2.2em][l]{$10^{#2}$}%
    }%
  }

  \resizebox{\linewidth}{!}{%
    \begin{tabular}{
      @{}l@{\hspace{8pt}}ccc@{\hspace{12pt}}ccc@{}
    }
      \toprule
      & \multicolumn{3}{c}{\textbf{CDR}}
      & \multicolumn{3}{c}{\textbf{MMSE}} \\
      \cmidrule(lr){2-4}
      \cmidrule(l){5-7}

      \textbf{Factor}
      & \textbf{OR [95\% CI]}
      & $\bm{p}$
      & $\bm{q}$
      & $\bm{\beta}$ \textbf{[95\% CI]}
      & $\bm{p}$
      & $\bm{q}$ \\
      \midrule

      Age
      & \orci{2.120}{1.823}{2.464}
      & \sci{1.33}{-22}
      & \sci{6.64}{-22}
      & \betaci{-0.849}{-0.979}{-0.692}
      & \sci{3.95}{-31}
      & \sci{1.97}{-30} \\

      Education
      & \orci{0.646}{0.533}{0.783}
      & \sci{8.09}{-6}
      & \sci{4.05}{-5}
      & \betaci{+0.406}{0.275}{0.574}
      & \sci{1.07}{-7}
      & \sci{5.35}{-7} \\

      APOE $\varepsilon4$
      & \orci{6.224}{3.623}{10.690}
      & \sci{3.48}{-11}
      & \sci{1.74}{-10}
      & \betaci{-1.584}{-2.562}{-0.844}
      & \sci{3.03}{-4}
      & \sci{7.57}{-4} \\

      Race
      & \orci{0.737}{0.506}{1.073}
      & \sci{1.12}{-1}
      & \sci{1.86}{-1}
      & \betaci{-0.426}{-0.735}{-0.089}
      & \sci{9.82}{-3}
      & \sci{1.64}{-2} \\

      SES
      & \orci{0.933}{0.290}{3.009}
      & \sci{9.08}{-1}
      & \sci{9.08}{-1}
      & \betaci{-1.825}{-3.426}{-0.559}
      & \sci{1.26}{-2}
      & \sci{3.15}{-2} \\

      ANXIETY
      & \orci{3.077}{1.438}{6.585}
      & \sci{3.78}{-3}
      & \sci{7.55}{-3}
      & \betaci{-0.246}{-0.740}{0.176}
      & \sci{2.92}{-1}
      & \sci{3.69}{-1} \\

      OCD
      & \orci{0.595}{0.065}{5.461}
      & \sci{6.46}{-1}
      & \sci{8.62}{-1}
      & \betaci{+0.894}{0.215}{1.638}
      & \sci{1.38}{-2}
      & \sci{2.55}{-2} \\

      ELAT
      & \orci{4.078}{1.688}{9.855}
      & \sci{1.79}{-3}
      & \sci{4.30}{-3}
      & \betaci{-0.108}{-1.258}{0.651}
      & \sci{8.24}{-1}
      & \sci{8.24}{-1} \\

      FOCLDEF
      & \orci{3.185}{1.121}{9.047}
      & \sci{2.96}{-2}
      & \sci{5.08}{-2}
      & \betaci{-1.934}{-3.714}{-0.822}
      & \sci{8.77}{-3}
      & \sci{1.75}{-2} \\

      \bottomrule
    \end{tabular}%
  }

  \endgroup
\end{table}

DeepQuest operationalized this question in OASIS-3 as a paired comparison of clinical impairment status and continuous cognitive performance within the same visit. The analysis included 1,376 participants with valid CDR and MMSE records. Clinical impairment was defined as CDRTOT $>0$, and MMSE represented continuous cognitive performance. Logistic regression was used for the CDR outcome, with odds ratios (OR) and marginal risk differences reported, while MMSE was modeled using censored regression to account for the 30 point ceiling. All models adjusted for age, sex, education, socioeconomic status, and study time.

DeepQuest first identified factors associated with both outcomes. A one standard deviation increase in age was associated with higher clinical impairment risk, with an OR of 2.120, and a 0.849 point decrease in MMSE. A one-standard-deviation increase in education was associated with lower odds of impairment (OR = 0.646) and higher MMSE scores ($\beta$ = 0.406). APOE $\varepsilon4$ homozygosity was also associated with higher impairment risk, with an OR of 6.224, and a 1.584 point decrease in MMSE. These results indicate that the two outcomes share several risk factors.

Several factors showed different patterns of statistical significance across outcomes. Race and socioeconomic status were primarily associated with MMSE, with effects of -0.426 and -1.825, respectively, but showed no significant association with CDR defined clinical impairment. Anxiety and elation were primarily associated with clinical impairment, with ORs of 3.077 and 4.078, but not with MMSE. In contrast, obsessive-compulsive disorder was significantly associated with higher MMSE, whereas focal neurological deficits were primarily associated with lower MMSE.

DeepQuest reproduced the finding that binary cognitive impairment status and continuous cognitive performance share some associated factors while also exhibiting distinct association patterns, with some factors showing opposite point estimate directions. The original study used clinical diagnoses of dementia and MCI as the binary outcome, whereas DeepQuest used CDRTOT $>0$ and censored regression for MMSE. These results therefore provide cross setting validation across different populations, outcome definitions, and analytical settings, showing that DeepQuest can recover both shared and outcome specific patterns in cognitive phenotypes.

\end{document}